\documentclass[lettersize,journal]{IEEEtran}
\usepackage{amsmath,amsfonts}
\usepackage{algorithmic}
\usepackage{algorithm}
\usepackage{array}
\usepackage[caption=false,font=normalsize,labelfont=sf,textfont=sf]{subfig}
\usepackage{textcomp}
\usepackage{stfloats}
\usepackage{url}
\usepackage{verbatim}
\usepackage{graphicx}
\usepackage{cite}
\usepackage{subfig}
\usepackage{booktabs}
\usepackage{booktabs}
\usepackage{multirow}
\usepackage{threeparttable}
\usepackage{adjustbox}

\newcommand{\best}[1]{\textbf{#1}}
\newcommand{\second}[1]{\underline{#1}}

\newtheorem{definition}{Definition}

\newtheorem{remark}{Remark}

\begin{document}

\title{Regional Explanations via\\Causal Sufficiency and Necessity}

\author{Xuexin Chen, Peng Liang, Zijian Li, Zhiyong Lin, and Ruichu Cai$^\star$

\thanks{(Corresponding author: Ruichu Cai.)}

\IEEEcompsocitemizethanks{
		\IEEEcompsocthanksitem Xuexin Chen is with the School of Artificial Intelligence, Guangdong Polytechnic Normal University, Guangzhou 510665, China.
		E-mail: im.chenxuexin@gmail.com
		\IEEEcompsocthanksitem Peng Liang is with the School of Artificial Intelligence, Guangdong Polytechnic Normal University, Guangzhou 510665, China. 
		E-mail: liangpeng@gpnu.edu.cn
		\IEEEcompsocthanksitem Zijian Li is with the 
         Mohamed bin Zayed University of Artificial Intelligence, Masdar City, Abu Dhabi, United Arab Emirates. 
		E-mail: leizigin@gmail.com
		\IEEEcompsocthanksitem Zhiyong Lin is with the School of Artificial Intelligence, Guangdong Polytechnic Normal University, Guangzhou 510665, China. 
        E-mail: zylin@gpnu.edu.cn
        \IEEEcompsocthanksitem Ruichu Cai is with the School of Computer Science and Technology, Guangdong University of Technology, Guangzhou 510006, China. 
		E-mail: cairuichu@gmail.com
}
}


\maketitle

\begin{abstract}
Model explainability is essential for understanding and trusting machine learning models. 
Existing explainable AI methods often explain predictions through feature importance, counterfactual explanations, or rules. 
However, a region-level characterization of when and only when a prediction behavior arises remains less explored. 
This paper proposes Causal Sufficient and Necessary Regional Explanations (SNRE), a framework that learns an input region $A$ and output region $B$ such that membership in $A$ is both sufficient and necessary for the model output to fall in $B$. 
Motivated by the classical Probability of Necessity and Sufficiency (PNS), we formulate a region-level PNS measure through stochastic interventions and derive a differentiable finite-sample estimator for optimization. 
SNRE parameterizes the input-output region pair with explicit and interpretable algebraic region families, together with a learnable feature mask, balancing expressiveness and interpretability. 
Experiments demonstrate that SNRE learns region pairs with strong sufficiency-necessity performance, robust explanation behavior, and practical utility for model analysis.
\end{abstract}


\section{Introduction}
\begin{figure}[!t]
\centering
\includegraphics[width=\linewidth]{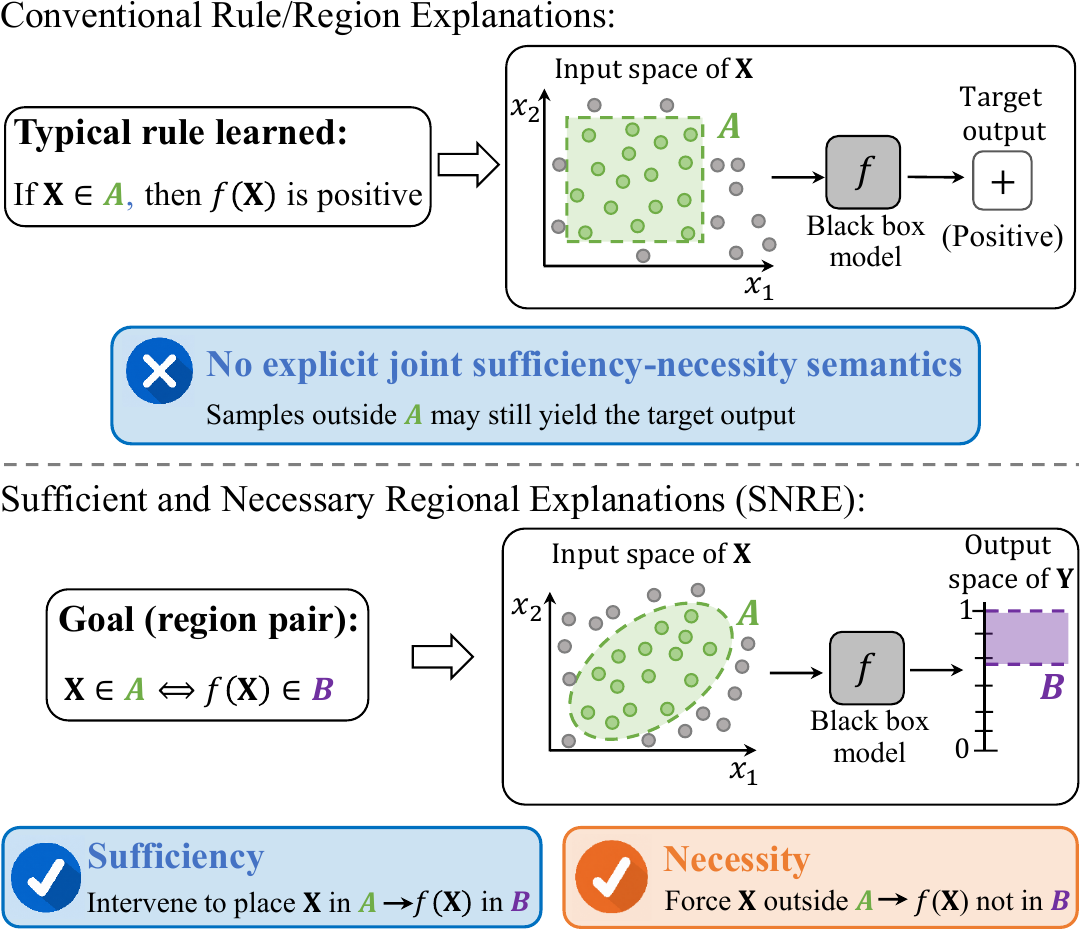}
\caption{
Conceptual comparison between conventional rule/region explanations and SNRE (Sufficient and Necessary Regional Explanations). Conventional rule/region explanations identify an input-side region $A$ associated with a target prediction, yet samples outside $A$ may yield the same output. 
SNRE learns an input-output region pair $(A,B)$ by contrasting in-region and out-of-region interventions, capturing both $A \to B$ and $\bar A \to \bar B$.
}
\label{fig:intro}
\end{figure}

Deep 
neural networks (DNNs) have achieved remarkable success across a wide range of domains, yet their highly non-linear and complex architectures make it difficult to understand why specific predictions are made.
This lack of transparency has motivated the rapid development of Explainable AI (XAI)\cite{zhao2023diml,chormai2024disentangled,kim2024does}, which seeks to make model behavior and predictions interpretable to human users\cite{rudin2019stop,guidotti2018survey,10475563}. 
Broadly, the XAI literature explains model behavior from multiple perspectives, 
including which input factors contribute to a prediction\cite{liu2026feature,nazir2025survey}, 
how the prediction can be changed by minimally modifying a given input\cite{verma2024counterfactual},
and what interpretable rules are associated with a target prediction behavior\cite{van2021evaluating}.

Beyond these established efforts, a fundamental but largely unexplored question in XAI is: can we identify an input-output region pair that exhibits a rigorous causal relationship? 
As illustrated in Fig.~\ref{fig:intro}, rule-based explanations\cite{mastromichalakis2024rule,souza2022decision} are the closest existing form to such a regional view, since a rule antecedent naturally defines an input-side condition.  
However, such regions typically specify where a target prediction may hold, but not where the same prediction becomes unlikely. 
This leaves the sufficiency-necessity structure of the prediction behavior unexplained. 
Formally, given a trained model $f$, we seek to discover a region $A$ in the input space and a target region $B$ in the output space: 
intervening to place the input within $A$ induces outputs in $B$, while forcing the input outside $A$ prevents outputs from falling in $B$. 
Such sufficient-and-necessary region pairs reveal when and only when a prediction behavior arises, 
and capture stable bidirectional regularities in the model's input-output mapping. 
Such a region-level structure provides a principled basis for downstream model analysis, such as examining changes in learned prediction regions after fine-tuning and guiding informative data selection.

To formalize such region-level causal relationships, the classical measure of causal sufficiency and necessity in causal inference, the \emph{Probability of Necessity and Sufficiency} (PNS) \cite{Pearl2009-PEACMR-3}, provides a natural theoretical foundation. However, extending this perspective to explanations in the form of continuous regions for DNNs face two fundamental challenges. 
First, classical PNS is defined for binary causes and outcomes, while regional explanations concern events $\mathbf X$$\in$$A$ and $f(\mathbf X)$$\in$$B$. This raises a basic question: how to define and estimate sufficiency-necessity strength for continuous input-output regions. 
Existing PNS-based XAI methods~\cite{chen2024feature,cai2025probability} largely bypass this difficulty by evaluating selected features or masks, without directly learning continuous causal regions in both input and output spaces.
Second, how to design an expressive but interpretable boundary representation for non-linear input/output region that reduces the geometric complexity  while preserving the attractive coverage property of necessity and sufficiency.

We propose \emph{Sufficient and Necessary Regional Explanations} (SNRE), a framework that learns input-output region pairs with high sufficiency-necessity strength by optimizing a region-level PNS objective motivated by classical causal sufficiency and necessity. 
We first formulate a region-level PNS measure for input and output regions, motivated by the core semantics of classical PNS. The key requirement is that the intervention distributions cover both the region and its complement, rather than collapsing to a narrow subset. 
This is naturally realized by two complementary \emph{stochastic interventions}~\cite{Pearl2009-PEACMR-3}, which draw in-region and out-of-region inputs from the data distribution. 
Based on this formulation, we derive a finite-sample estimator of region-level PNS and further construct a differentiable relaxation for optimization. SNRE then parameterizes the input-output region pair with explicit and interpretable algebraic region families, together with a learnable soft feature mask for sparse feature selection. By jointly optimizing the region parameters and the mask, SNRE learns both the regional boundaries and the sparse features that characterize the sufficiency-necessity relation.
The learned region pairs can further serve as explicit structural summaries of model behavior, enabling downstream analyses such as tracking regional changes after model fine-tuning and guiding informative data selection. 
Finally, we conduct extensive experiments to evaluate the sufficiency-necessity performance, structural rationality, robustness, and practicality of SNRE.

The main contributions of our work are as follows: 
1) We formulate a region-level PNS measure for input-output regions, motivated by classical causal sufficiency and necessity, and instantiate it through complementary stochastic interventions with a finite-sample estimator and differentiable relaxation. 
2) We propose an end-to-end SNRE framework that learns sufficient-and-necessary input-output region pairs by optimizing the region-level PNS objective under explicit and interpretable algebraic region parameterizations with a learnable soft feature mask. 
3) We conduct comprehensive experiments to validate the explanation quality, structural rationality, robustness, and practical utility of SNRE.

\section{Related work}

\subsection{Regional Explanations}

Regional explanations describe model behavior over input-space regions.
A common realization is rule-based explanation, where rule antecedents naturally define an input region $A$.
Early work by Cohen~\cite{cohen1995fast} develops efficient rule induction for classification, and Wang and Rudin~\cite{wang2017bayesian} introduce Bayesian rule sets to represent predictions with compact logical conditions.
Lakkaraju et al.~\cite{lakkaraju2016interpretable} learn interpretable decision sets optimized for accuracy, conciseness, and low overlap.
Yang et al.~\cite{pmlr-v70-yang17h} learn scalable Bayesian rule lists from pre-mined candidate rules, while Wang and Rudin~\cite{pmlr-v38-wang15a} impose a monotonic falling probability structure along ordered rules.
Angelino et al.~\cite{angelino2018learning} learn certifiably optimal rule lists, and Proen{\c c}a and van Leeuwen~\cite{proencca2020interpretable} select compact probabilistic rule lists under the minimum description length principle.
Pellegrina et al.~\cite{pellegrina2024scalable} further improve the scalability of rule-list learning on large datasets. Tree-based methods also induce regional explanations by recursively partitioning the input space.
Loh~\cite{loh2011classification} reviews decision-tree methods whose leaves naturally define input regions, and Good et al.~\cite{good2023feature} combine sparse feature transformation learning with differentiable decision-tree fitting to obtain compact trees.
Recent differentiable and logic-based learners further improve rule learning through neural optimization or logical constraints.
Wang et al.~\cite{wang2021scalable} learn non-fuzzy logical rules via differentiable relaxation, Qiao et al.~\cite{qiao2021learning} incorporate logical rule components into neural architectures, and Ghosh et al.~\cite{ghosh2022efficient} learn sparse propositional rules through an incremental framework.

However, these methods mainly define input-side regions associated with target predictions, rather than directly learning continuous input-output region pairs.
In contrast, SNRE learns such region pairs based on sufficiency-necessity principles.

\subsection{PNS-based Explanations}

Probability of Necessity and Sufficiency (PNS)~\cite{pearl2022probabilities,tian2000probabilities} provides a causal perspective for post-hoc explanations by evaluating whether an explanatory factor is sufficient to preserve a prediction and necessary for producing it.
Existing studies instantiate this principle over different explanatory units.
Watson et al.~\cite{watson2021local} identify minimal feature subsets with necessity and sufficiency guarantees.
Balkir et al.~\cite{balkir2022necessity} assess token-level text explanations through necessity and sufficiency.
Tan et al.~\cite{tan2022learning} evaluate graph subgraphs via factual and counterfactual reasoning. 
Galhotra et al.~\cite{galhotra2021explaining} compute attribute-level probabilistic contrastive counterfactual scores for local, global, and contextual explanations.
Cai et al.~\cite{cai2025probability} optimize PNS lower bounds for graph explanations, while Chen et al.~\cite{chen2024feature} estimate feature-level necessity and sufficiency probabilities through intervention-based attribution.


However, these methods mainly evaluate sufficiency and necessity over features, tokens, subgraphs, attributes, or masks, rather than directly learning continuous regions in both the input and output spaces.
In contrast, SNRE directly learns interpretable input-output region pairs under a region-level sufficiency-necessity formulation.

\section{Background and Problem Formulation}

We start with some basic notation. 
We use an uppercase letter, such as $X$, to denote a random variable, and a lowercase letter, such as $x$, to denote its realization when the context is clear. 
A bold uppercase letter, such as $\mathbf X$, denotes a random vector, while a bold lowercase letter, such as $\mathbf x$, denotes a vector in the corresponding value space.
For a random vector $\mathbf X$, we write $\mathrm{Val}_{\mathbf X}$ for its value space, so that $\mathbf x\in \mathrm{Val}_{\mathbf X}$. 
A fixed trained predictor is denoted by $f:\mathrm{Val}_{\mathbf X}\to \mathrm{Val}_{\mathbf Y}$. 
We use $A\subseteq \mathrm{Val}_{\mathbf X}$ and $B\subseteq \mathrm{Val}_{\mathbf Y}$ to denote input and output regions, respectively.

\subsection{Causal Sufficiency and Necessity}
In the context of model interpretability, rather than modeling underlying real-world causal mechanisms, we adopt existing model-based causal perspective\cite{geiger2025causal}, where the input variables $\mathbf X$ are treated as causes that without parents, while the model output $\mathbf Y$ as the effect. 
Interventions correspond to actively setting the input to specific values or regions and observing the resulting model output.
Accordingly, sufficiency and necessity are defined with respect to the behavior of a fixed trained predictor under such input interventions. 


We consider explanations in terms of regions in the input and output spaces.
Let $A \subset \mathrm{Val}_{\mathbf X}$ denote an input region and $B \subset \mathrm{Val}_{\mathbf Y}$ denote an output region of interest.

At the region level, we distinguish three different causal relationships between the events $\mathbf X \!\in\! A$ and $\mathbf Y \!\in\! B$. 
1) Sufficient but unnecessary cause $A$: the input region $A$ is said to be a \emph{sufficient but unnecessary cause} of the output region $B$ if intervening to place the input within $A$ is enough to induce an output in $B$, but $B$ may still be obtained even when the input is outside $A$.  
2) necessary but
insufficient cause $A$: $A$ is a \emph{necessary but
insufficient cause} of $B$ if, without placing the input within $A$, the output would not fall in $B$, but intervening to place the input within $A$ alone does not guarantee an output in $B$. 
3) sufficient and necessary cause $A$: 
$A$ is a \emph{sufficient and necessary cause} of $B$, if placing the input within $A$ induces an output in $B$, and without placing the input within $A$, an output in $B$ would not be obtained.

These notions capture a causal regularity at the region level: the occurrence of the event $\mathbf X \in A$ explains the occurrence of the event $\mathbf Y \in B$ by being both required and adequate.

\subsection{Problem Formulation}  
The region level notions of sufficiency and necessity above are conceptually appealing,
but turning them into a practical objective requires a quantitative measure that can be
computed and optimized for predictors.

In this work, we consider a trained deterministic predictor
$f: \mathrm{Val}_{\mathbf X}$$\to$$\mathrm{Val}_{\mathbf Y}$ and aim to
estimate an input region $A \subset \mathrm{Val}_{\mathbf X}$ and an output
region $B \subset \mathrm{Val}_{\mathbf Y}$ such that membership in $A$
exhibits a strong sufficiency-necessity relation with membership in $B$,
as quantified by a PNS-based measure. 
We focus on the common setting where $f$ is fixed after training and differentiable with respect to its input.  
Our goal is to develop a computable and optimizable formulation
that enables learning such sufficient-and-necessary region pairs from finite data. 
Here, ``deterministic'' means that a fixed input $\mathbf x$ uniquely determines $f(\mathbf x)$. 
This includes standard sigmoid or softmax classifiers, whose probability or score outputs are deterministic functions of the input. 
Models with stochastic prediction, such as Bayesian neural networks \cite{blundell2015weight} with sampled parameters, are beyond the scope of this work.

\begin{remark}[Relation to decision regions]
For a fixed predictor and output event $B$ (e.g., $B=[0.5,1]$),  the exact \emph{decision region} can be written as $\mathcal S^B = \{\mathbf x\in \mathrm{Val}_{\mathbf X}\mid f(\mathbf x)\in B\}$,
which satisfies $\mathbf x\in \mathcal S^B \Leftrightarrow f(\mathbf x)\in B$. 
However, for neural predictors, $\mathcal S^B$ is usually implicit, highly non-linear, and difficult to interpret\cite{montufar2014number}. 
Therefore, our goal is not to recover the full decision region, 
but to extract an explicit and geometrically interpretable region pair $(A,B)$ that satisfies the sufficiency-necessity relation with high probability.
\end{remark}

\section{Region-level PNS via Stochastic Interventions}
\label{sec:cpns}

We begin by revisiting the classical \emph{Probability of Necessity and Sufficiency} (PNS) in causal inference, and use its core semantics to motivate a region-level sufficiency-necessity measure for input-output regions through \emph{stochastic interventions}.

\subsection{Motivating Region-level PNS from Classical PNS}
In causal inference, the PNS~\cite{Pearl2009-PEACMR-3,pearl2022probabilities,tian2000probabilities}, introduced by Pearl, provides a principled way to quantify whether a cause is both sufficient and necessary for an outcome.
Classical PNS is defined for binary causes and binary outcomes, where events correspond to whether a variable takes value $1$ or $0$.
Let $X \in \{0,1\}$ be a binary cause and $Y \in \{0,1\}$ be a binary outcome.
Using the potential outcomes notation~\cite{rubin1974estimating}, let $Y_x$ denote the outcome that would be observed under the hard intervention $do(X=x)$.
The classical PNS is defined as
\begin{equation}
\label{eq:bin_pns}
\mathrm{PNS}
:=
 P\big(
Y_1 = 1 \;\wedge\; Y_0 = 0
\big).
\end{equation}
This quantity measures the probability that setting $X=1$ would make the outcome occur, while setting $X=0$ would prevent the outcome from occurring.

A key observation is that PNS compares the model’s output behavior under two complementary interventions:
one that forces the input to satisfy a specified condition, and another that forces it to violate that condition.
In the binary setting, ``satisfying the condition'' corresponds to setting $X$$=$$1$, while ``violating it'' corresponds to setting $X$$=$$0$, i.e., $do(X$$=$$1)$ versus $do(X$$=$$0)$. 

\begin{figure}[!t]
\centering
\includegraphics[width=0.5\textwidth]{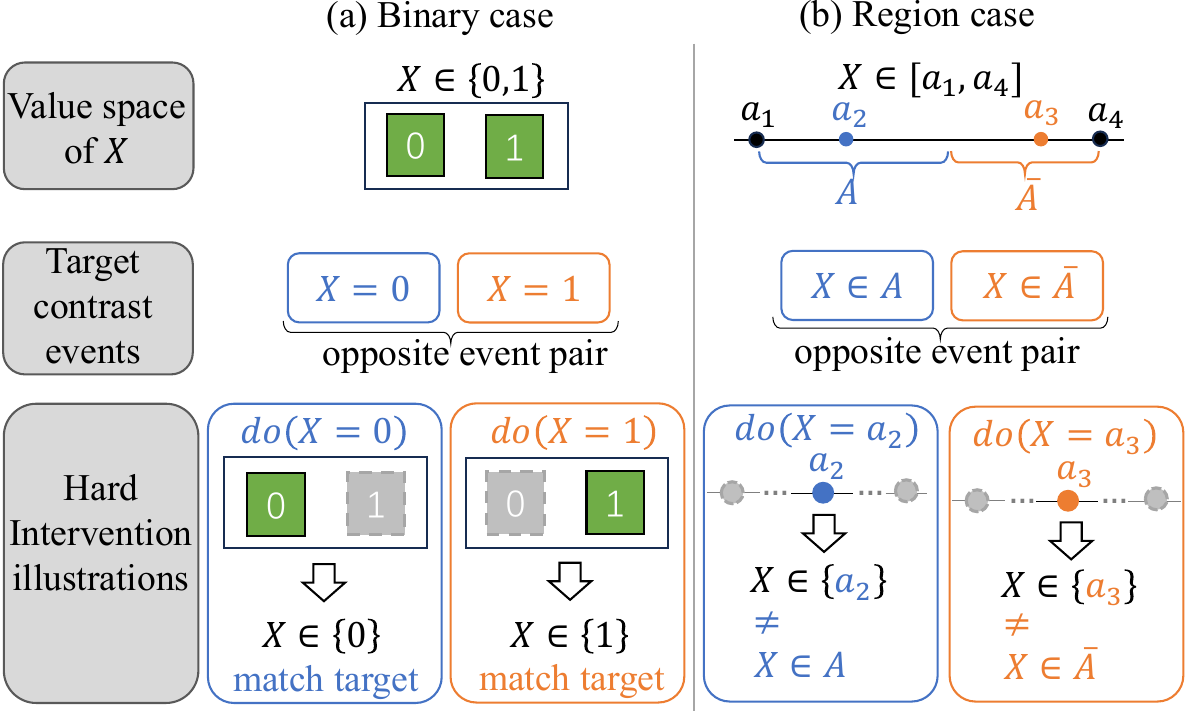}
\caption{Illustration of why hard interventions fail for region-level events. In the binary case, hard interventions exactly match the two opposite target events. In the region case, hard interventions collapse the value space to singleton events and therefore cannot realize the contrast between $X\in A$ and $X \in \bar A$.
}
\label{fig:int}
\end{figure}

However, this mechanism does not directly extend beyond binary variables.
When the input space is not binary, causal statements are naturally expressed as \emph{events} of set membership, e.g., the event $\mathbf X$$\in$$A$ versus $\mathbf X$$\notin$$A$, rather than equality to a single value.
An event-level analogue of PNS thus requires contrasting interventions that enforce $\mathbf X$$\in$$A$ and $\mathbf X$$\notin$$A$.
The standard hard intervention form cannot directly realize such event-level interventions. 
A hard intervention $do(\mathbf X=\mathbf x)$ fixes the input to a single value, which degenerates the induced interventional distribution to a point. 
As a result, it cannot represent the typical model behavior over inputs \emph{inside} versus \emph{outside} a set (or region) $A$, as illustrated in Fig.~\ref{fig:int}.

\subsection{Stochastic Interventions for Regional Events}
To preserve sufficiency-necessity semantics at the event level, interventions must induce non-degenerate coverage over both the region of interest and its complement within a common domain. 
We therefore adopt two complementary \emph{stochastic intervention}~\cite{Pearl2009-PEACMR-3} regimes defined with respect to the input distribution. 
Specifically, let $P(\mathbf X)$ denote the distribution over the input space, and let $\mathcal D \subseteq \mathrm{Val}_{\mathbf{X}}$ be a \emph{reference input domain} on which input events are defined. A canonical choice is the support of the input distribution, 
\begin{equation}
    \mathcal{D} := \mathrm{supp}(P(\mathbf{X})).
\end{equation}
The domain $\mathcal D$ is not restricted to purely continuous spaces and may include discrete or mixed components.
In practice, $\mathcal D$ is instantiated implicitly by the observed samples
$\{\mathbf x_i\}_{i=1}^n$.

For a measurable set $A\! \subseteq$$\mathcal D$ (interpreted as a region when $\mathcal D$ is continuous), define its complement as $\bar A$$:=$$\mathcal D$$\setminus$$A$.
Conditioning $P(\mathbf{X})$ on $A$ and $\bar A$ yields two restricted-support distributions:
\begin{equation}
    P^A(\mathbf X):=P(\mathbf X \mid\! \mathbf X \in A), 
    P^{\bar A}(\mathbf X):=P(\mathbf X \mid\! \mathbf X \in \bar A).
\end{equation}
These distributions define two complementary stochastic intervention schemes, denoted by
$\pi(\mathbf X;P^A)$ and $\pi(\mathbf X;P^{\bar A})$.
Each scheme independently samples a value $\mathbf x$ from the corresponding conditional distribution and applies the hard intervention $do(\mathbf X$$=$$\mathbf x)$, reflecting two separate counterfactual worlds as in the classical PNS formulation.
By construction, the supports of $P^A$ and $P^{\bar A}$ form a partition of the domain $\mathcal D$, and provide non-degenerate coverage of both $A$ and $\bar A$.

\subsection{Region-Level PNS and Estimation}

We now define region-level PNS under the complementary stochastic intervention policies
$\pi(\mathbf X;P^A)$ and $\pi(\mathbf X;P^{\bar A})$.
Let $\mathbf Y_{\pi(\mathbf X;P^A)}$ denote the model output under policy $\pi(\mathbf X;P^A)$, and define $\mathbf Y_{\pi(\mathbf X;P^{\bar A})}$ analogously.
The output space $\mathrm{Val}_{\mathbf Y}$ is given by the value space of the trained model, e.g., $\mathbb R^{|\mathbf Y|}$ or $(0,1)$.
For any measurable output set $B $$\subset$$\mathrm{Val}_{\mathbf Y}$, with complement $\bar B := \mathrm{Val}_{\mathbf Y} \setminus B$, we consider the output events $\mathbf Y$$\in$$B$ and $\mathbf Y$$\in$$\bar B$.
Recall that the input domain is $\mathcal D := \mathrm{supp}(P(\mathbf X))$, where input events and their complements are defined.

\begin{definition}[Region-level PNS]
\label{def:cont_pns}
For a measurable input set $A \!\subset\! \mathcal D$ and a measurable output set
$B \!\subset\! \mathrm{Val}_\mathbf Y$, the region-level probability of sufficiency and necessity
is defined as
\begin{equation}
\label{equ:cpns}
    \mathrm{PNS}(A,B)
    :=
    P\big(
    \mathbf Y_{\pi(\mathbf X;P^A)} \in B
    \;\wedge\;
    \mathbf Y_{\pi(\mathbf X;P^{\bar A})} \in \bar B
    \big).
\end{equation}
\end{definition}
This quantity measures the sufficiency-necessity strength between the input event $\mathbf X \in A$ and the output event $\mathbf Y \in B$ under complementary stochastic interventions.

Since the predictor $f$ is deterministic and the input variables $\mathbf X$ are treated as \emph{exogenous} under the model-based causal perspective~\cite{geiger2025causal}, the output is a deterministic function of the intervened input.
The two intervention policies are sampled independently, so the joint event in \eqref{equ:cpns} can be estimated by separately sampling from the in-region and out-of-region intervention distributions.
Given $m$ i.i.d.\ samples
$\{\mathbf x_i^{\mathrm{in}}\}_{i=1}^m$$\sim$$P(\mathbf X$$\mid$$\mathbf X$$
$$\in$$A$$)$
and
$\{\mathbf x_i^{\mathrm{out}}\}_{i=1}^m$$\sim$$P(\mathbf X \mid$$\mathbf X$$\in$$\bar A$$)$, 
we estimate $\mathrm{PNS}(A,B)$ via
\begin{equation}
\label{eq:pns_mc}
\widehat{\mathrm{PNS}}(A,B)
\!=\!
\left(\frac{1}{m}\!\sum_{i=1}^{m}\!\mathbb I\!\big[f(\mathbf x_i^{\mathrm{in}})\!\in\!B\big]\!\right)\!
\left(\frac{1}{m}\!\sum_{i=1}^{m}\!\mathbb I\!\big[f(\mathbf x_i^{\mathrm{out}})\in\!\bar B\big]\!\right)\!.
\end{equation}
As $m\to\infty$, $\widehat{\mathrm{PNS}}(A,B)$ converges to $\mathrm{PNS}(A,B)$ whenever both conditional distributions are well defined, i.e., both $A$ and $\bar A$ have nonzero probability mass under $P(\mathbf X)$.
A formal justification is provided in Appendix~\ref{app:pns_mc}. 
In implementation, we require both $A$ and $\bar A$ to contain empirical samples to avoid trivial degeneration.

\begin{figure}[!t]
\centering
\includegraphics[width=2.5in]{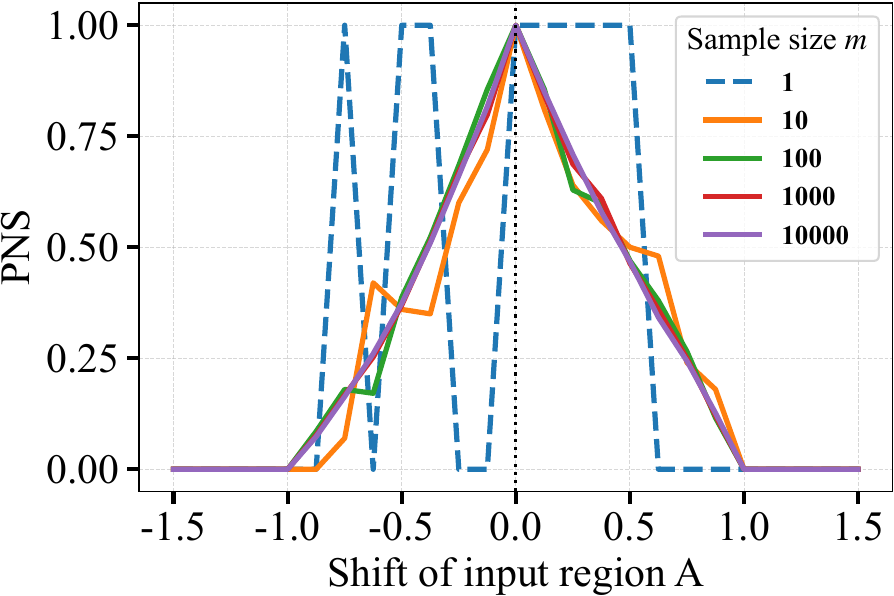}
\caption{Monte Carlo estimates of $\mathrm{PNS}(A,B^\star)$ as the continuous input region $A$ is shifted away from an optimal reference region $A^\star$ under different sample sizes $m$.}
\label{fig:cpns}
\end{figure}

Fig.~\ref{fig:cpns} visualizes $\widehat{\mathrm{PNS}}(A,B^\star)$ when the continuous input region $A$ is progressively shifted away from a reference region $A^\star$, while the output region is fixed to $B^\star$.
Here $(A^\star,B^\star)$ is a constructed region pair that attains the highest PNS in this controlled setting.
Each curve corresponds to a different sample size $m$ used in \eqref{eq:pns_mc}.
When $m=1$, the estimator uses only one sampled intervention per side, and even incorrect regions can achieve spuriously high PNS values, reflecting that such sparse intervention samples cannot represent the typical model behavior over a region and therefore cannot reliably distinguish $A^\star$ from incorrect alternatives.
As $m$ increases, the estimator increasingly captures the model's typical output behavior over the conditional distributions on $A$ and $\bar A$: $\widehat{\mathrm{PNS}}(A,B^\star)$ peaks sharply near $A^\star$ and decays monotonically as $A$ moves away from it.
This demonstrates that region-level PNS can meaningfully quantify sufficiency-necessity strength over continuous input-output regions.

\begin{remark}[Model-based intervention and identification]
Unlike classical causal identification, SNRE does not aim to recover unobserved potential outcomes of an unknown real-world structural causal model from observational data. 
Instead, it treats the fixed trained predictor as the causal system~\cite{geiger2025causal}, whose responses under input interventions can be directly evaluated. 
The remaining problem is therefore finite-sample estimation of the stochastic intervention distributions, rather than identification of unknown counterfactual responses. 

\end{remark}

\section{SNRE: Sufficient and Necessary Regional Explanations}
\begin{figure*}[!t]
\centering
\includegraphics[width=\linewidth]{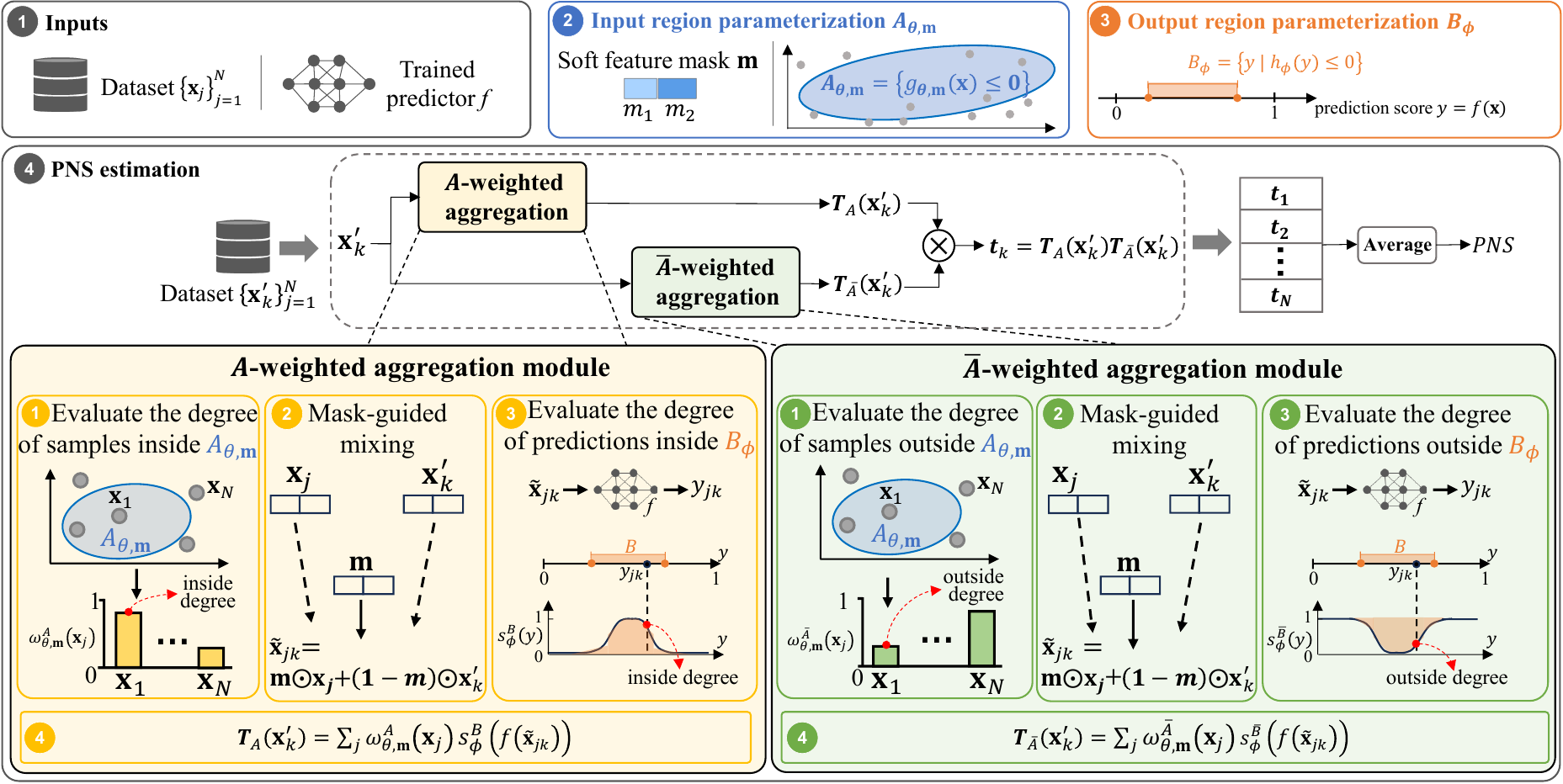}
\caption{
SNRE framework for learning sufficient-and-necessary input-output region pairs. 
SNRE first parameterizes the input region $A_{\theta,\mathbf m}$ and output region $B_{\phi}$, and then estimates PNS through outer marginalization and two parallel weighted aggregation modules. 
The $A$-weighted module evaluates how strongly inputs inside $A_{\theta,\mathbf m}$ induce predictions inside $B_{\phi}$, while the $\bar A$-weighted module evaluates how strongly inputs outside $A_{\theta,\mathbf m}$ induce predictions outside $B_{\phi}$. 
The final PNS objective is obtained by averaging the product of the two aggregation terms over the training samples used for marginalization.}
\label{fig:model}
\end{figure*}

Building on the region-level PNS formulation in Section~\ref{sec:cpns},
we now present \emph{Sufficient and Necessary Regional Explanations} (SNRE),
an end-to-end framework for learning input-output region pairs $(A,B)$ with high sufficiency-necessity strength. 
As illustrated in Fig.~\ref{fig:model}, SNRE takes a fixed trained predictor $f$ and training samples as inputs, and learns an input region $A_{\theta,\mathbf m}$ and an output region $B_{\phi}$ by optimizing a differentiable relaxation of region-level PNS. 
SNRE parameterizes the input and output regions with explicit learnable region functions, and introduces a soft feature mask $\mathbf m$ to select the input dimensions involved in the sufficiency-necessity relation. 
It then estimates region-level PNS through sample-based marginalization over the dimensions not emphasized by the mask and two parallel weighted aggregation modules: 
the $A$-weighted module estimates the in-region-to-$B_{\phi}$ term, while the $\bar A$-weighted module estimates the out-of-region-to-$\bar B_{\phi}$ term.
The product of these two aggregation scores yields a differentiable objective for jointly optimizing the region parameters and the feature mask.

\subsection{Region Parameterization and Feature Mask}
Given a trained predictor $f: \mathrm{Val}_{\mathbf X} \to \mathrm{Val}_{\mathbf Y}$ and input data, 
our goal is to estimate an input region $A \subset \mathcal D$ and an output region $B \subset \mathrm{Val}_\mathbf Y$ that maximize the region-level $\mathrm{PNS}(A,B)$, 
where $\mathcal D$ denotes the empirical input domain instantiated by the available dataset. 
Since direct optimization over all measurable sets is intractable, we  parameterize the regions $A$ and $B$ via differentiable functions. 

\subsubsection{Input region parameterization} 
We parameterize the boundary of the input region $A$ by a quadratic hypersurface. 
Let $g_{\theta}: \mathrm{Val}_{\mathbf X} \to \mathbb{R}$ be defined as
\[
g_{\theta}(\mathbf{x}) = \mathbf{x}^{\top} \mathbf{w} \mathbf{x} + \mathbf{v}^{\top} \mathbf{x} + b,
\]
where $\theta = (\mathbf{w}, \mathbf{v}, b)$ are learnable parameters, 
$\mathbf w\in\mathbb R^{d\times d}$ is a symmetric matrix, 
$\mathbf v\in\mathbb R^d$, $b\in\mathbb R$, and 
$d=\mathrm{dim}(\mathrm{Val}_{\mathbf X})$. 
The corresponding input region is
\begin{equation}
A_{\theta} = \{\mathbf{x} \in \mathcal{D} \mid g_{\theta}(\mathbf{x}) \leq 0\}.    
\end{equation}

The boundary $g_\theta(\mathbf x)=0$ forms a quadratic hypersurface, which offers a favorable balance between expressivity and interpretability: it can represent a rich family of shapes, such as ellipsoids, hyperboloids, and paraboloids, while its algebraic form permits analytic reasoning about geometric properties such as centers, principal axes, and curvature. 
We empirically validate the effectiveness of this input-region parameterization in Section~\ref{sec:structural_rationality}.

\subsubsection{Output region parameterization}
In general, the output region $B$ can be parameterized by any explicit and interpretable region family that matches the structure of the output space. 
For higher-dimensional outputs, $B$ may also be represented by a quadratic hypersurface, analogously to the input region. 
In this work, we focus on classification settings, where the model output is typically low-dimensional, such as a class probability or a vector of class scores. 
The target behavior is therefore naturally associated with a compact confidence region, so a lower-complexity parameterization is sufficient.

When the output region needs to be learned, we use the following hyperspherical form:
\begin{equation}
\label{eq:h_out_quad}
h_{\phi}(\mathbf y)=(\mathbf y-\boldsymbol{\mu})^\top(\mathbf y-\boldsymbol{\mu})-r^2,
\end{equation}
where $\phi=(\boldsymbol{\mu},r)$ are learnable parameters, and $r>0$ is enforced by $r=\mathrm{softplus}(\rho)$ for a free parameter $\rho$. 
The induced output region is
\begin{equation}
\label{eq:B_phi}
B_{\phi}=\{\mathbf y\in \mathrm{Val}_{\mathbf Y}\mid h_{\phi}(\mathbf y)\le 0\}.
\end{equation}
Note that the target output event can also be specified in advance, in which case $B$ can be fixed or only partially learned.

\subsubsection{Feature selection via a learnable mask} 
The quadratic form of $g_\theta$ naturally admits a feature-selection mechanism. 
The linear coefficients $\mathbf{v}$ in $g_\theta$ capture the main effect of each input dimension, while the quadratic term $\mathbf{w}$ encodes pairwise interactions. To identify the subset of input features that are most relevant for the sufficiency-necessity relation, we introduce a continuous, learnable mask $\mathbf{m} \in (0, 1)^d$. Specifically, we parameterize $\mathbf{m}$ via a sigmoid function $\sigma$ over free parameters $\tilde{\mathbf{m}} \in \mathbb R^d$, i.e., $\mathbf m = \sigma(\tilde{\mathbf{m}})$. The mask is then incorporated into the quadratic region definition by gating each input dimension: 
\begin{equation}\label{equ:gw}
\begin{aligned}
&g_{\theta, \mathbf{m}}(\mathbf{x}) = \mathbf{x}^{\top} \tilde{\mathbf{w}} \mathbf{x} + \tilde{\mathbf{v}}^{\top} \mathbf{x} + b,\\
&\tilde{\mathbf{w}} = \mathbf{u} \mathbf{w} \mathbf{u}, \tilde{\mathbf{v}} = \mathbf{u}\mathbf{v, }\mathbf{u} = \operatorname{diag}(\sqrt{m_1}, \dots, \sqrt{m_d}).    
\end{aligned}
\end{equation}
The use of $\sqrt{m_j}$ rather than $m_j$ itself avoids overly suppressing the quadratic term and leads to more stable gradients. When $m_j \approx 0$, the $j$-th feature contributes neither to the linear nor to the quadratic part of $g_{\theta,\mathbf{m}}$, effectively removing it from the region definition. During training $\mathbf{m}$ is kept continuous to allow gradient-based optimization; at inference we threshold $\mathbf{m}$ (e.g., with a cutoff at 0.5) to obtain a binary mask $\tilde{\mathbf{m}} \in\{0,1\}^d$, which defines the selected feature subset. Thus, the input region is now jointly parameterized by $(\theta, \mathbf{m})$, and we write 
\begin{equation}
A_{\theta, \mathbf{m}} = \{\mathbf{x} \in \mathcal{D} \mid g_{\theta,\mathbf{m}}(\mathbf{x}) \le 0\}.
\end{equation}
Here, the mask $\mathbf m$ specifies the input dimensions used to define the regional event $A$. 
In the explicit region function in \eqref{equ:gw}, this selection is implemented by gating the corresponding dimensions. 
As described later in \eqref{equ:final_obj}--\eqref{eq:mix_mask}, the remaining dimensions are treated as background variables and marginalized when evaluating the intervention effect through the fixed predictor.

\subsection{Differentiable PNS Objective and Optimization}
Maximizing the exact $\mathrm{PNS}(A_{\theta,\mathbf m}, B_{\phi})$ is challenging because Definition~\ref{def:cont_pns} involves hard event judgments and conditional input
distributions that are not available in closed form. 
We therefore construct a differentiable Monte Carlo estimator via a sigmoid relaxation.

Given a dataset $\{\mathbf x_i\}_{i=1}^n$, we introduce a soft membership score
for the input event $\mathbf X\in A_{\theta, \mathbf{m}}$: 
\begin{equation}
\label{eq:soft_in}
s^A_{\theta,\mathbf{m}}(\mathbf x)
=\sigma\!\big(-g_{\theta,\mathbf m}(\mathbf x)/\tau\big), 
\end{equation}
and a corresponding soft membership score for the output event $\mathbf Y\in B_\phi$:
\begin{equation}
\label{eq:soft_out}
s^B_\phi(\mathbf y)
=\sigma\!\big(-h_\phi(\mathbf y)/\tau\big),
\end{equation}
where $\sigma(\cdot)$ denotes the sigmoid function and $\tau$$>$$0$ is a temperature parameter.
These quantities provide smooth approximations to the indicator functions
$\mathbb I(\mathbf x$$\in$$A_{\theta,\mathbf{m}})$ and $\mathbb I(\mathbf{y}\in B_\phi)$, respectively. Accordingly, for the complementary events $\mathbf X$$\in$$\bar A_{\theta, \mathbf{m}}$ and $\mathbf Y$$\in$$\bar B_\phi$,
we use
\begin{equation}
\label{eq:soft_comp}
s^{\bar A}_{\theta,\mathbf m}(\mathbf x):= 1-s^A_{\theta,\mathbf m}(\mathbf x),\quad
s^{\bar B}_\phi(\mathbf y):= 1-s^B_\phi(\mathbf y).
\end{equation}

Substituting the above relaxations into the Monte Carlo estimator of $\mathrm{PNS}(A_{\theta, \mathbf{m}}, B_\phi)$ in \eqref{eq:pns_mc}, yields the following differentiable objective:
\begin{equation}
\label{equ:final_obj}
\begin{aligned}
&\widehat{\mathrm{PNS}}(A_{\theta, \mathbf{m}}, B_\phi)=\\
&\frac{1}{N}\sum_{k=1}^N\!\sum_{j=1}^N \omega^A_{\theta,\mathbf{m}}(\mathbf x_j)\, s^B_\phi\!\big(f(\tilde{\mathbf x}_{jk})\big)
\sum_{j=1}^N \omega^{\bar A}_{\theta, \mathbf{m}}(\mathbf x_j)\, s^{\bar B}_\phi\!\big(f(\tilde{\mathbf x}_{jk})\big),  
\end{aligned}
\end{equation}
where $\tilde{\mathbf x}_{jk}$ denotes the mixed intervention input constructed from two training samples $\mathbf x_j$ and $\mathbf x_k$ under the soft mask $\mathbf m$:
\begin{equation}
\label{eq:mix_mask}
\tilde{\mathbf x}_{jk}
= \mathbf m \odot \mathbf x_j + (1-\mathbf m)\odot \mathbf x_k.    
\end{equation}
Intuitively, dimensions emphasized by the mask are mainly taken from $\mathbf x_j$ to enforce the input event, while dimensions not emphasized by the mask are supplied by $\mathbf x_k$ and averaged over the dataset for sample-based marginalization.
The weights $\omega^A_{\phi,\mathbf m}(\cdot)$ and $\omega^{\bar A}_{\theta, \mathbf{m}}(\cdot)$ are normalized coefficients used to approximate the conditional expectations under the stochastic intervention policies $\pi(\mathbf{X}; P^A)$ and $\pi(\mathbf{X}; P^{\bar A})$ by empirical averages over the dataset.
They are defined as
\begin{equation}
\label{eq:omega}
\begin{aligned}
&\omega^A_{\theta, \mathbf{m}}(\mathbf x_j)
=\frac{s^A_{\theta, \mathbf{m}}(\mathbf x_j)}
{\sum_{\ell=1}^N s^A_{\theta, \mathbf{m}}(\mathbf x_\ell)},\\
&\omega^{\bar A}_{\theta, \mathbf{m}}(\mathbf x_j)
=\frac{s^{\bar A}_{\theta, \mathbf m}(\mathbf x_j)}
{\sum_{\ell=1}^N s^{\bar A}_{\theta,\mathbf m}(\mathbf x_\ell)}.
\end{aligned}
\end{equation}
A detailed derivation of \eqref{equ:final_obj} is provided in Appendix~\ref{app:pns_derivation}.

\subsection{Extracting Sufficient-and-Necessary Regional Explanations}
After optimization converges, we binarize the learned mask to obtain
$\mathbf m^*\in\{0,1\}^d$ by thresholding: $m^*_j=1$ if $m_j>0.5$ and $m^*_j=0$ otherwise. This yields the selected feature subset
$\mathcal I^*=\{j: m^*_j=1\}$. 
The resulting input and output regions are
\begin{equation}
\begin{aligned}
A^* : = A_{\theta^*, \mathbf{m}^*} = \{\mathbf{x} \in \mathcal{D} \mid g_{\theta^*,\mathbf{m}^*}(\mathbf{x}) \leq 0\},\\
B^* := B_{\phi^*} = \{\mathbf y \in \mathrm{Val}_{\mathbf{Y}} \mid h_{\phi^*}(\mathbf{y}) \leq 0\}.   
\end{aligned}
\end{equation}
The pair $(A^*,B^*)$ constitutes a \emph{sufficient-and-necessary regional explanation} for
the model $f$ with respect to the selected features $\mathcal I^*$.  
It summarizes a regional sufficiency-necessity relation: inputs placed within $A^*$ are mapped to $B^*$ with high probability, while inputs placed outside $A^*$ are mapped outside $B^*$ with high probability.

\begin{table*}[!t]
\centering
\caption{Summary of datasets used in the experiments. All tasks are formulated as binary classification.}
\label{tab:dataset_summary}
\footnotesize
\setlength{\tabcolsep}{4pt}
\renewcommand{\arraystretch}{1.1}
\begin{tabular}{lll}
\toprule
Dataset & Task & Usage in Experiments \\
\midrule
German Credit\cite{hoffmanstatlog} & Predict good/bad credit risk & Region quality, active learning \\
GMSC\cite{GiveMeSomeCredit} & Predict serious delinquency within two years & Region quality, robustness \\
Mushroom\cite{mushroom_73} & Predict edible versus poisonous mushrooms & Region quality, mask robustness \\
Phishing Websites\cite{mohammad2014intelligent} & Predict phishing versus legitimate websites & Region quality, geometry ablation,  efficiency \\
Synthetic Circles & Two-dimensional concentric-circle classification & Toy visualization of SNRE optimization \\
Synthetic Blobs & Two-dimensional synthetic classification & Fine-tuning-induced explanation drift \\
\bottomrule
\end{tabular}
\end{table*}


\section{Experiments}

Building on the proposed region-level PNS formulation and the SNRE optimization framework, 
we evaluate SNRE on four real-world tabular datasets and two synthetic datasets
with a summary given in Table~\ref{tab:dataset_summary}. The experiments are organized around five questions.
1) Can a toy example provide intuition about the joint optimization of the input region $A$ and output region $B$ under the proposed region-level PNS objective? 
2) On real datasets, do the learned region pairs exhibit strong sufficiency and necessity on unseen test samples, and outperform existing PNS-based and rule-based explanation methods?
3) Are the proposed design choices structurally rational, particularly the quadratic input-region parameterization and the learnable feature mask? 
4) How robust and efficient is SNRE?
5) How practical are the region-pair explanations generated by SNRE?

For all experiments, we use the same post-hoc explanation setup.
We first train a predictor $f$ on the training split and keep it fixed during explanation learning.
SNRE and all baselines are trained on the same train/validation/test split.
SNRE learns a sufficient-and-necessary region pair $(A,B)$ with respect to the prediction behavior of $f$, where the input region $A$ and output region $B$ are parameterized separately and jointly optimized by maximizing the proposed sufficiency-necessity objective until convergence.
Each experiment is repeated five times, and we report the mean result.

We compare SNRE with 17 baselines, including 11 rule-based explanation methods and 6 PNS-based explanation methods.
For rule-based baselines, we train each method using the model output as the target rather than the ground-truth label. 
For PNS-based methods, since they typically return feature scores or feature subsets instead of continuous input-output region pairs, 
we convert their selected features into an input event $A$ using the same box post-processing protocol. Specifically, for a selected feature subset $\mathcal S$, we define
$A^{\mathcal S}=\{\mathbf x: l_j \le x_j \le u_j,\ \forall j\in \mathcal S\}$. 
After comparing several box construction strategies, we use a mean-std box, where the bounds are estimated from training samples satisfying the fixed output event $B$:
$l_j=\mu_j-\rho\sigma_j, u_j=\mu_j+\rho\sigma_j$, 
where $\mu_j$ and $\sigma_j$ are the empirical mean and standard deviation of feature $j$, and $\rho$ is selected on the validation split.

For more details on the baselines, metrics, model configurations and hyperparameters, please refer to Appendix~\ref{app:detail_exp}.

\subsection{Toy Example: Visualizing Joint Learning of Input-Output Region Pairs}
\label{sec:toy_example}
\begin{figure}[!t]
\centering
\includegraphics[width=\linewidth]{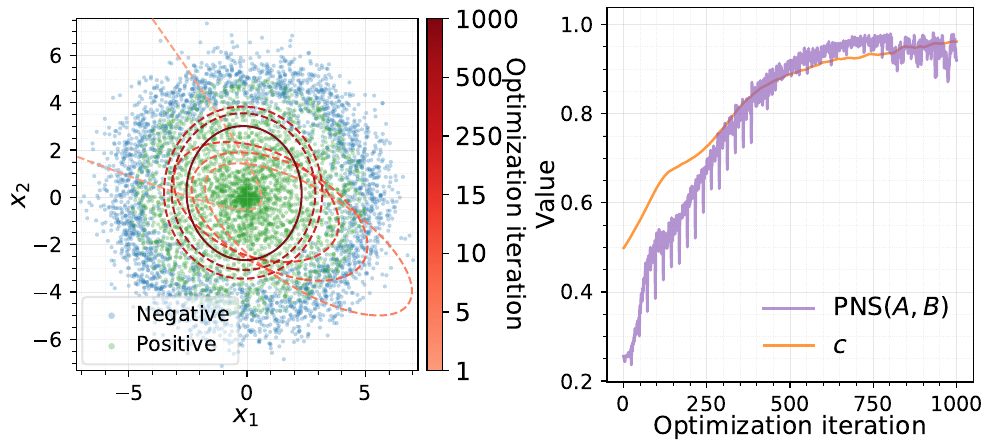}
\caption{Toy example for joint learning of an input-output region pair $(A,B)$ with SNRE. 
Left: snapshots of the learned input region $A$, represented by the quadratic boundary $g(\mathbf{x})=0$, across optimization iterations. 
Colors indicate snapshot iterations, with dashed contours for intermediate snapshots and a solid contour for the final one. 
Right: optimization trajectories of the region-level $\mathrm{PNS}(A,B)$ and the output region parameter $c$ defining $B=[c,1]$ over optimization iterations.}
\label{fig:snre_dynamics}
\end{figure}

We construct a simple binary-classification toy example to make the optimization behavior of SNRE visually tangible, focusing on learning a sufficient-and-necessary region pair $(A,B)$ associated with positive predictions. 
The data consist of two concentric circles in a 2D continuous input space. 
A neural network classifier is trained to separate the inner disk (positive class, shown in green) from the outer ring (negative class, shown in blue). 
SNRE is then run for 1,000 optimization steps to jointly learn an input region $A$, represented by a quadratic boundary, together with an output region $B$. 
Because the output is one-dimensional, the parameterization in \eqref{eq:h_out_quad} reduces to an interval on the real line, and since this experiment targets positive outputs, we use the restricted form $B=[c,1]$, where $c$ is a learnable lower endpoint.

In this toy setting, the sufficient-and-necessary region pair associated with the positive class is expected to align $A$ with the inner disk and to place $B$ near the upper end of the model's output distribution. 
Fig.~\ref{fig:snre_dynamics} (left) visualizes training snapshots of the learned input-region boundary.
At early iterations, the boundary (light dashed contours) is poorly aligned with the data geometry and often traverses the transition band between the two classes.
As optimization proceeds, the boundary progressively tightens around the inner disk, eventually (dark red and solid contour) forming a close fit that cleanly separates the high-confidence positive samples from the negative ones. 
The coupled dynamics on the output side are shown in Fig.~\ref{fig:snre_dynamics} (right). 
The lower bound $c$ of $B$ (orange solid line) rises steadily from about 0.5 to a value near 1.0, shifting $B$ toward the high-confidence regime. 
This shift is essential: to maintain high sufficiency, $A$ must avoid the ambiguous region where predictions are uncertain, and forcing $B$ to high probabilities naturally drives $A$ to contract onto the core of the positive class. 
Throughout this joint optimization of $(A, B)$, the $\mathrm{PNS}(A, B)$ objective (purple line) increases steadily and approaches $1$, indicating that the learned region pair becomes progressively more sufficient and necessary. 

\subsection{Quantitative Evaluation of Learned Sufficient-and-necessary Region Pairs}
\begin{table*}[!t]
\caption{Comparison of PNS, PN, and PS on four datasets. Results are reported as mean$\pm$std over five folds and rounded to three decimals. The best and second-best results in each column are highlighted in bold and underlined, respectively.}
\label{tab:main_results}
\centering
\footnotesize
\setlength{\tabcolsep}{3.2pt}
\renewcommand{\arraystretch}{1.08}
\begin{threeparttable}
\adjustbox{max width=\textwidth}{
\begin{tabular}{@{}lcccccccccccc@{}}
\toprule
\multirow{2}{*}{Method}
& \multicolumn{3}{c}{Mushroom}
& \multicolumn{3}{c}{GMSC}
& \multicolumn{3}{c}{Phishing Websites}
& \multicolumn{3}{c}{German Credit} \\
\cmidrule(lr){2-4}\cmidrule(lr){5-7}\cmidrule(lr){8-10}\cmidrule(lr){11-13}
& PNS & PN & PS & PNS & PN & PS & PNS & PN & PS & PNS & PN & PS \\
\midrule
IDS     
& 0.255$\pm$0.000 & 0.527$\pm$0.000 & 0.483$\pm$0.000
& 0.559$\pm$0.000 & 0.993$\pm$0.000 & 0.563$\pm$0.000
& 0.328$\pm$0.000 & 0.591$\pm$0.000 & 0.555$\pm$0.000
& 0.464$\pm$0.000 & 0.801$\pm$0.000 & 0.580$\pm$0.000 \\

SBRL    
& 0.254$\pm$0.000 & 0.526$\pm$0.000 & 0.483$\pm$0.000
& 0.573$\pm$0.000 & 0.996$\pm$0.000 & 0.575$\pm$0.000
& 0.442$\pm$0.000 & 0.759$\pm$0.000 & 0.582$\pm$0.000
& 0.447$\pm$0.000 & 0.718$\pm$0.000 & 0.622$\pm$0.000 \\

FRL     
& 0.256$\pm$0.000 & 0.529$\pm$0.004 & 0.483$\pm$0.000
& 0.573$\pm$0.000 & 0.996$\pm$0.000 & 0.575$\pm$0.000
& 0.328$\pm$0.000 & 0.591$\pm$0.000 & 0.555$\pm$0.000
& 0.383$\pm$0.000 & 0.758$\pm$0.000 & 0.506$\pm$0.000 \\

DT      
& 0.293$\pm$0.008 & 0.597$\pm$0.016 & 0.490$\pm$0.000
& 0.826$\pm$0.000 & 0.996$\pm$0.000 & 0.829$\pm$0.000
& 0.442$\pm$0.000 & 0.759$\pm$0.000 & 0.582$\pm$0.000
& 0.418$\pm$0.000 & 0.815$\pm$0.000 & 0.513$\pm$0.000 \\

FLDT    
& 0.547$\pm$0.156 & 0.813$\pm$0.125 & 0.662$\pm$0.096
& \second{0.881}$\pm$0.006 & \second{0.998}$\pm$0.000 & \underline{0.883}$\pm$0.006
& 0.725$\pm$0.029 & 0.897$\pm$0.034 & 0.809$\pm$0.041
& \second{0.720}$\pm$0.034 & 0.844$\pm$0.038 & 0.853$\pm$0.039 \\

RRL     
& \second{0.966}$\pm$0.030 & \second{0.992}$\pm$0.006 & \second{0.974}$\pm$0.025
& 0.677$\pm$0.057 & 0.994$\pm$0.001 & 0.681$\pm$0.058
& \second{0.893}$\pm$0.017 & \second{0.949}$\pm$0.016 & \second{0.941}$\pm$0.007
& 0.665$\pm$0.022 & 0.822$\pm$0.023 & 0.810$\pm$0.033 \\

CORELS  
& 0.268$\pm$0.003 & 0.553$\pm$0.004 & 0.486$\pm$0.000
& 0.559$\pm$0.000 & 0.993$\pm$0.000 & 0.563$\pm$0.000
& 0.442$\pm$0.000 & 0.759$\pm$0.000 & 0.582$\pm$0.000
& 0.296$\pm$0.000 & 0.549$\pm$0.000 & 0.540$\pm$0.000 \\

SamRuLe 
& 0.268$\pm$0.000 & 0.553$\pm$0.000 & 0.486$\pm$0.000
& 0.559$\pm$0.000 & 0.993$\pm$0.000 & 0.563$\pm$0.000
& 0.442$\pm$0.000 & 0.759$\pm$0.000 & 0.582$\pm$0.000
& 0.441$\pm$0.006 & 0.724$\pm$0.008 & 0.610$\pm$0.006 \\

MDL-RL  
& 0.274$\pm$0.004 & 0.562$\pm$0.005 & 0.488$\pm$0.000
& 0.573$\pm$0.000 & 0.996$\pm$0.000 & 0.575$\pm$0.000
& 0.487$\pm$0.000 & 0.792$\pm$0.000 & 0.614$\pm$0.000
& 0.415$\pm$0.000 & 0.718$\pm$0.000 & 0.578$\pm$0.000 \\

DRNet   
& 0.509$\pm$0.230 & 0.761$\pm$0.158 & 0.644$\pm$0.158
& 0.598$\pm$0.348 & 0.978$\pm$0.002 & 0.611$\pm$0.356
& 0.536$\pm$0.032 & 0.540$\pm$0.035 & 0.993$\pm$0.006
& 0.582$\pm$0.002 & 0.590$\pm$0.002 & \best{0.985}$\pm$0.001 \\

IMLI    
& 0.264$\pm$0.000 & 0.545$\pm$0.001 & 0.484$\pm$0.000
& 0.541$\pm$0.000 & 0.979$\pm$0.000 & 0.553$\pm$0.000
& 0.645$\pm$0.000 & 0.899$\pm$0.000 & 0.718$\pm$0.001
& 0.527$\pm$0.009 & \best{0.869}$\pm$0.012 & 0.606$\pm$0.009 \\
\midrule

FANS 
& 0.380$\pm$0.132 & 0.648$\pm$0.092 & 0.573$\pm$0.116
& 0.538$\pm$0.067 & \second{0.998}$\pm$0.001 & 0.539$\pm$0.068
& 0.679$\pm$0.062 & 0.887$\pm$0.035 & 0.766$\pm$0.051
& 0.546$\pm$0.017 & 0.692$\pm$0.065 & 0.797$\pm$0.101 \\

LEWIS  
& 0.503$\pm$0.000 & 0.674$\pm$0.000 & 0.746$\pm$0.000
& 0.347$\pm$0.000 & \best{0.999}$\pm$0.000 & 0.347$\pm$0.000
& 0.675$\pm$0.000 & 0.874$\pm$0.000 & 0.773$\pm$0.000
& 0.458$\pm$0.000 & 0.557$\pm$0.000 & 0.823$\pm$0.000\\

LENS   
& 0.250$\pm$0.000 & 0.482$\pm$0.000 & 0.519$\pm$0.000
& 0.454$\pm$0.000 & \second{0.998}$\pm$0.000 & 0.455$\pm$0.000
& 0.467$\pm$0.087 & 0.784$\pm$0.082 & 0.591$\pm$0.059
& 0.507$\pm$0.042 & 0.710$\pm$0.064 & 0.722$\pm$0.121 \\

NSEG  
& 0.268$\pm$0.010 & 0.488$\pm$0.002 & 0.550$\pm$0.021
& 0.316$\pm$0.083 & \second{0.998}$\pm$0.000 & 0.317$\pm$0.084
& 0.597$\pm$0.001 & 0.854$\pm$0.002 & 0.699$\pm$0.002
& 0.467$\pm$0.004 & 0.724$\pm$0.023 & 0.645$\pm$0.022 \\

CF$^2$   
& 0.261$\pm$0.000 & 0.487$\pm$0.000 & 0.536$\pm$0.000
& 0.247$\pm$0.110 &  \second{0.998}$\pm$0.000 & 0.247$\pm$0.110
& 0.730$\pm$0.005 & 0.882$\pm$0.000 & 0.827$\pm$0.006
& 0.470$\pm$0.013 & 0.753$\pm$0.008 & 0.624$\pm$0.011 \\

NSE4T   
& 0.836$\pm$0.031 & 0.907$\pm$0.011 & 0.921$\pm$0.023
& 0.454$\pm$0.030 & \second{0.998}$\pm$0.011 & 0.455$\pm$0.023
& 0.718$\pm$0.034 & 0.884$\pm$0.022 & 0.812$\pm$0.020
& 0.481$\pm$0.040 & 0.752$\pm$0.012 & 0.639$\pm$0.045 \\

\midrule
SNRE    
& \best{0.988}$\pm$0.004 & \best{0.999}$\pm$0.001 & \best{0.990}$\pm$0.004
& \best{0.982}$\pm$0.002 & 0.982$\pm$0.002 & \best{1.000}$\pm$0.000
& \best{0.950}$\pm$0.006 & \best{0.984}$\pm$0.002 & \best{0.966}$\pm$0.005
& \best{0.750}$\pm$0.020 & \second{0.861}$\pm$0.059 & \second{0.882}$\pm$0.063 \\
\bottomrule
\end{tabular}
}
\end{threeparttable}
\end{table*}

We evaluate whether the learned explanations preserve strong sufficiency and necessity on unseen test data. 
For each method, we derive an input region $A$ and measure its relation to a fixed, high-confidence output interval $B$ ($B=[0.8,1]$ for Mushroom, German Credit, and Phishing Websites, and $B=[0.5,1]$ for GMSC). 
For SNRE, $(A,B)$is its learned region pair. 
For rule-based baselines (upper block of Table~\ref{tab:main_results}), $A$ is the rule antecedent. 
For PNS-based explanation methods (middle block of Table~\ref{tab:main_results}), which usually output feature subsets or feature scores rather than regions, we post-process their results into a comparable input region A. 
Specifically, we use a mean-std bounding box, which was found to be the most effective post-processing strategy after extensive testing.

As shown in the table~\ref{tab:main_results}, SNRE achieves the highest PNS on all four datasets, indicating the best overall generalization performance of the learned explanations on the test set. On Mushroom, GMSC, and Phishing Websites, SNRE not only leads in PNS, but also attains the best or near-best PS and PN. On German Credit, although SNRE does not achieve the highest value in either PS or PN individually, the higher scores of other methods are more one-sided. 
For example, IMLI yields a higher PN but a lower PS, while DRNet attains a  high PS but suffers from a lower PN. 
In contrast, SNRE achieves a better balance between the two, and therefore still obtains the highest PNS overall. 
Therefore, the advantage of SNRE does not lie in maximizing a single metric in isolation, but in learning a more generalizable and better coordinated sufficient-and-necessary explanation.

\subsection{Structural Rationality Analysis of Region Parameterization}
\label{sec:structural_rationality}

\subsubsection{Input region parameterization}

\begin{figure*}[!t]
\centering
\includegraphics[width=\linewidth]{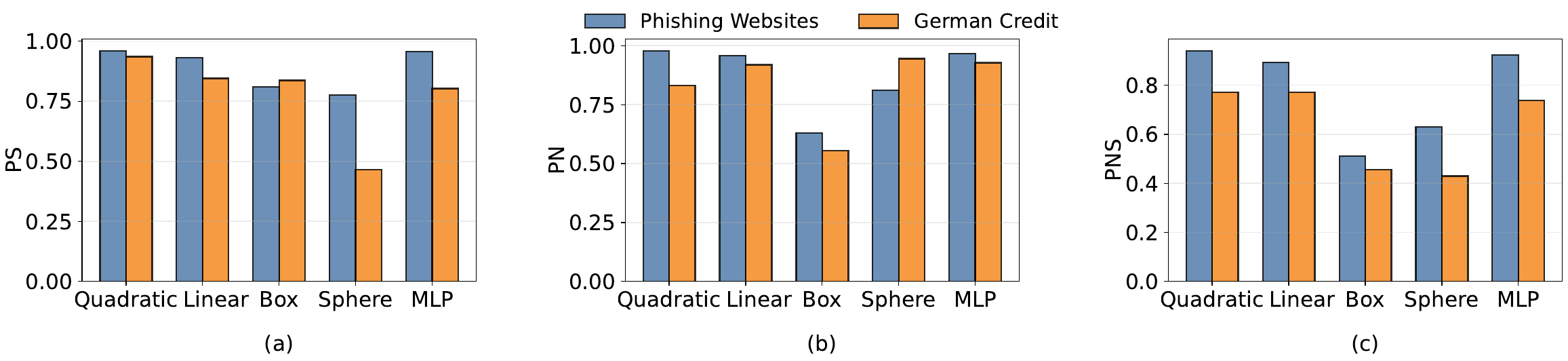}
\caption{Ablation on input-region parameterization. We compare five region families, including linear, box-shaped, spherical, quadratic, and MLP-based implicit boundaries, on Phishing Websites and German Credit. The results are reported in terms of PN, PS and PNS. Quadratic parameterization achieves a favorable balance between expressive power and explanation quality across datasets.}
\label{fig:geometry_ablation_pns}
\end{figure*}

\begin{figure*}[!t]
\centering
\includegraphics[width=\linewidth]{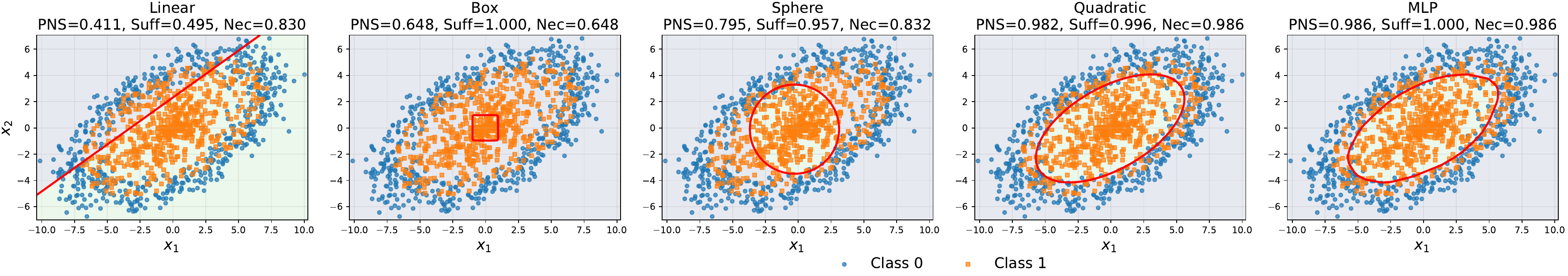}
\caption{Comparison of input region shapes from different parameterizations. A 2D example shows that linear, box, and spherical shapes are too rigid to fit the data, while an MLP learns a complex, irregular boundary. The quadratic hypersurface forms a regular ellipse, best balancing geometric expressiveness and interpretability.}
\label{fig:geometry_ablation_vis}
\end{figure*}

To validate the structural rationality of quadratic hypersurfaces, we compare five input-region families: linear, box, sphere, quadratic, and MLP-based implicit boundaries. As shown in Fig.~\ref{fig:geometry_ablation_pns}, the quadratic parameterization achieves the best or near-best PNS on both datasets. On Phishing Websites, it delivers the highest PS, PN, and PNS simultaneously, indicating that its geometric flexibility is sufficient to capture the target sufficient-and-necessary region. On German Credit, although it is not uniformly best on each individual term, it provides the most balanced trade-off between PS and PN, and thus remains optimal overall in PNS. In contrast, linear, box, and spherical regions are geometrically constrained and therefore struggle to jointly preserve coverage and exclusivity, leading to a clearer sufficiency-necessity mismatch. 

Fig.~\ref{fig:geometry_ablation_vis} further explains this effect geometrically. On the rotated elliptical structure, linear boundaries are too rigid to align with the true region, while box and sphere parameterizations cannot capture rotation and anisotropy. By contrast, the quadratic boundary naturally matches the underlying elliptical geometry, which enables high sufficiency and necessity simultaneously. Although the MLP parameterization achieves a comparable fit, its boundary is implicit and lacks the explicit geometric semantics of a quadratic form. These results show that quadratic hypersurfaces provide a more favorable structural trade-off between expressive power and interpretability for sufficient-and-necessary regional explanations.

\subsubsection{Feature Mask}
\begin{figure}[!t]
\centering
\includegraphics[width=\linewidth]{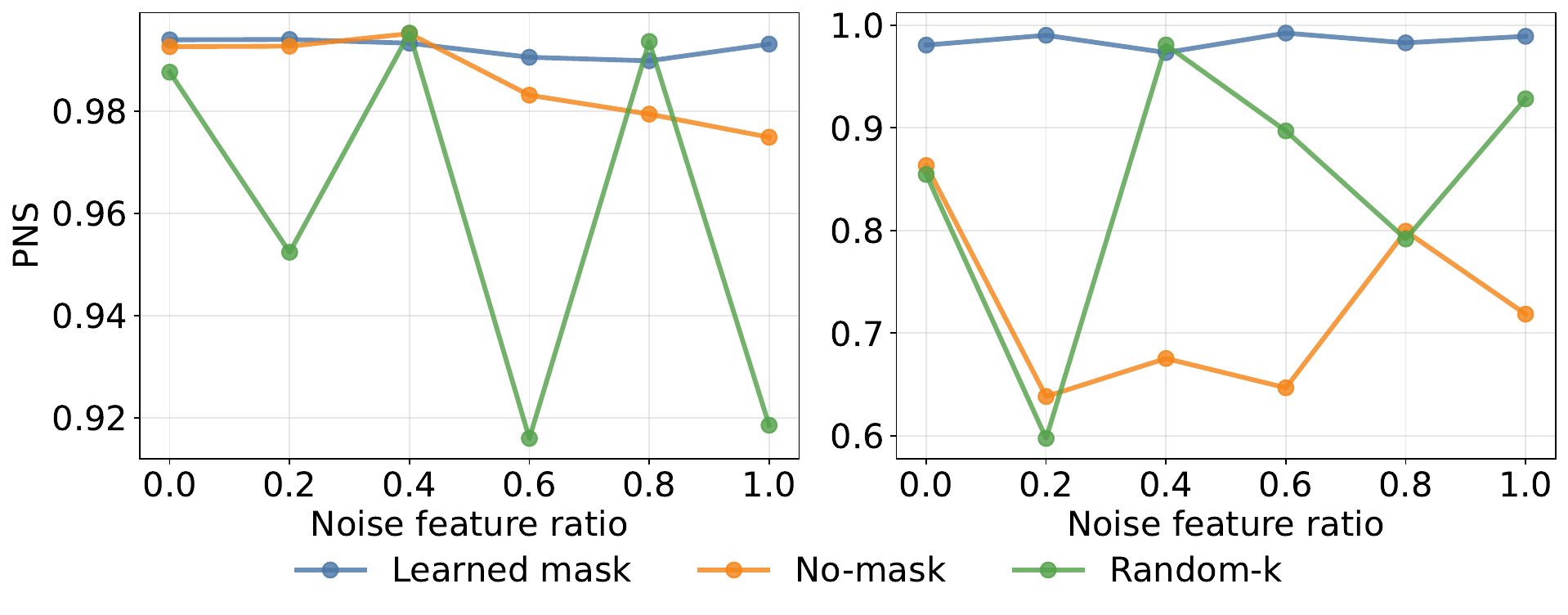}
\caption{PNS under different feature-mask strategies with increasing noise-feature ratios.}
\label{fig:feat_mask}
\end{figure}

We next investigates the rationality of learning a feature mask within the SNRE optimization process. Specifically, we design a comparative experiment with three settings: \textit{Learned mask} (the proposed SNRE method that jointly learns feature importance), \textit{No-mask} (standard training without masking), and \textit{Random-k} (randomly selecting kfeatures as a baseline). We evaluate the stability of the PNS metric under varying ratios of noise features on two datasets.

As shown in Fig.~\ref{fig:feat_mask}, the \textit{Learned mask} strategy consistently maintains high PNS scores under both low and high noise ratios in both datasets, demonstrating strong robustness. In contrast, the \textit{No-mask} and \textit{Random-k} strategies exhibit significant fluctuations and sharp declines in PNS as the noise ratio increases. 
This indicates that learning a mask allows the model to ignore irrelevant noisy features during optimization, preventing the degradation of the decision boundary for the core mechanism, 
while the other two methods show much poorer PNS stability as the noise ratio increases.

\subsection{Robustness analysis of SNRE explanations}
\begin{figure}[!t]
\centering
\includegraphics[width=\linewidth]{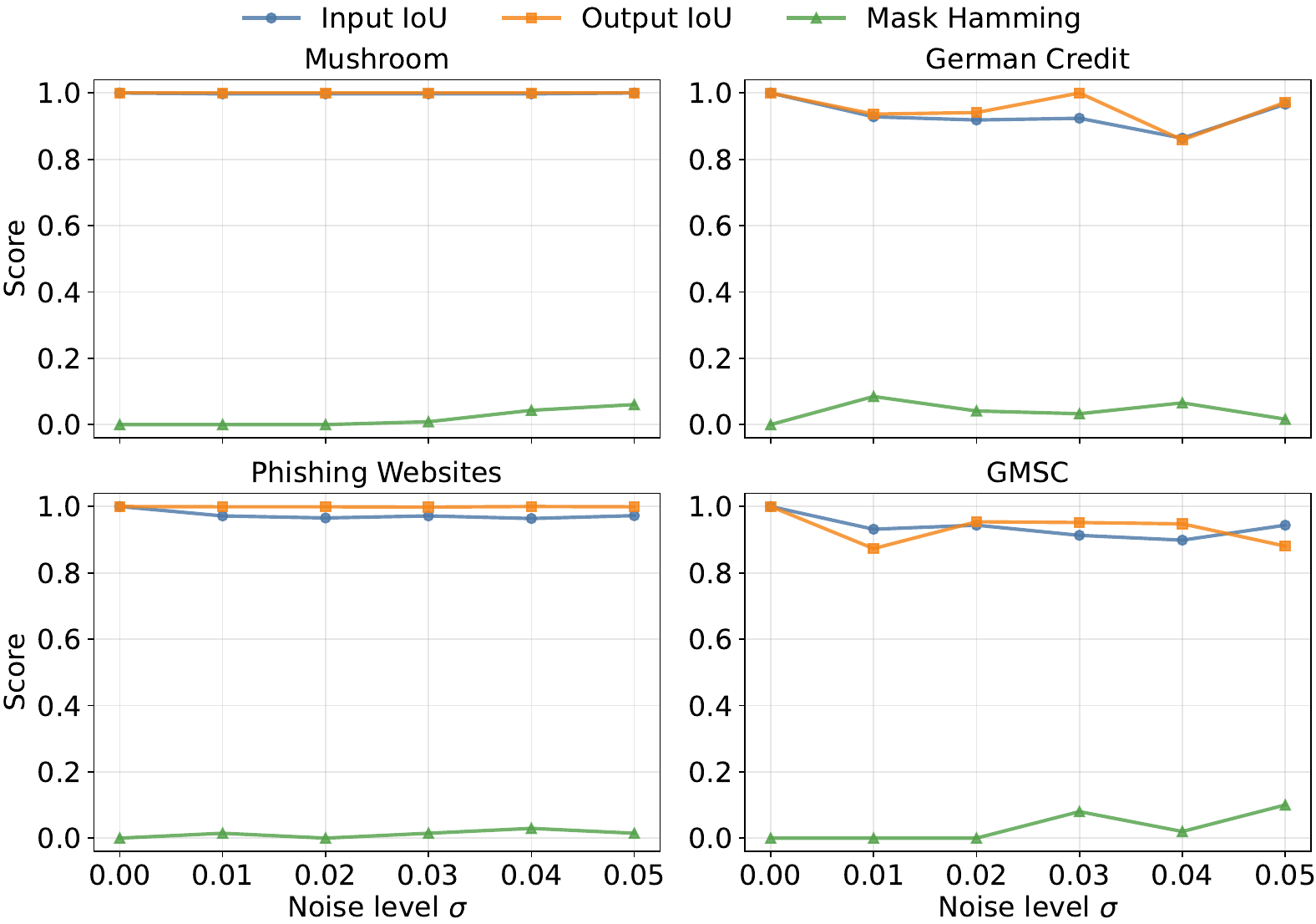}
\caption{Robustness analysis of SNRE explanations under increasing training input noise across four datasets, measured by Input IoU, Output IoU, and Mask Hamming.}
\label{fig:explanation_stability}
\end{figure}

This section investigates the robustness of the explanations produced by SNRE under input perturbations. Specifically, we fix the same black-box model and data split, take the explainer trained on the clean training set as the reference explainer, and then retrain the explainer after injecting Gaussian noise with increasing magnitude into the training inputs only. The resulting explanations are compared against the reference one using three consistency measures: Input IoU, Output IoU, and Mask Hamming. Let $\mathbf{a}^{\mathrm{ref}}$ and $\mathbf{a}^{\mathrm{cur}}$ denote the binary membership vectors of the input region produced by the reference and perturbed explainers, respectively, and let $\mathbf{b}^{\mathrm{ref}}$ and $\mathbf{b}^{\mathrm{cur}}$ denote the corresponding binary membership vectors of the output region. Both input-region and output-region consistency are measured by the intersection-over-union:
\[
\mathrm{IoU}(\mathbf{u}, \mathbf{v}) = \frac{|\mathbf{u} \cap \mathbf{v}|}{|\mathbf{u} \cup \mathbf{v}|}.
\]
Accordingly, Input IoU is computed from $(\mathbf{a}^{\mathrm{ref}}, \mathbf{a}^{\mathrm{cur}})$, and Output IoU is computed from $(\mathbf{b}^{\mathrm{ref}}, \mathbf{b}^{\mathrm{cur}})$. In addition, let $\mathbf{m}^{\mathrm{ref}}, \mathbf{m}^{\mathrm{cur}} \in \{0,1\}^{d}$ denote the hard feature masks of the two explainers. Their discrepancy is measured by the normalized Hamming distance:
\[
\mathrm{Hamming}(\mathbf{m}^{\mathrm{ref}}, \mathbf{m}^{\mathrm{cur}})
=
\frac{1}{d}\sum_{j=1}^{d}\mathbf{1}\!\left(m^{\mathrm{ref}}_{j} \neq m^{\mathrm{cur}}_{j}\right).
\]
Hence, higher Input/Output IoU and lower Mask Hamming indicate better explanation stability under noise.

As shown in Fig.~\ref{fig:explanation_stability}, SNRE exhibits strong stability across all four datasets. On Mushroom, both Input IoU and Output IoU remain almost at 1.0 across different noise levels, while Mask Hamming stays close to 0, indicating nearly unchanged explanations under perturbation. On German Credit, Phishing Websites, and GMSC, although moderate fluctuations can be observed, both IoU scores remain consistently high overall and the mask discrepancy stays small. These results suggest that the input region, output region, and selected explanatory features learned by SNRE do not drift substantially under noisy training inputs, demonstrating good robustness and consistency of the resulting explanations.

\subsection{Model Size and Efficiency}
\begin{table}[htbp]
  \centering
  \caption{Model size and efficiency analysis on the PhishingWebsites dataset.}
    \begin{tabular}{lcrrr}
    \toprule
    Method & \multicolumn{1}{c}{\# Params} & \multicolumn{1}{c}{Training (min)} & \multicolumn{1}{c}{Inference (ms)} & \multicolumn{1}{c}{PNS} \\
    \midrule
    IDS   &   --    &   0.4169    &   0.0023   &  0.326\\
    SBRL  &    --   &    0.5078    &     0.0055   &  0.444\\
    FRL   &   --    &  \textbf{0.4121}   &     0.0045   &  0.326\\
    DT   &   --    &    2.6360  &  0.0242    &  0.444\\
    FLDT   &   --    &   55.3380   &   0.0252     &  0.76\\
    RRL   &    87,509   &    19.2264   &   0.3256  &  \underline{0.895}\\
    CORELS &    --   &   \underline{0.4196}   &   0.0052   &  0.444\\
    SamRule &    --   &   2.3039  &   0.0046    &  0.444\\
    MDL-RL &    --   &   0.5028   &   0.0147  &  0.490\\
    DRNet &    \textbf{6,240}  &  3.9330   &    0.1180    &  0.\\
    IMLI   &    --   &   6.9016    &   0.0029   &  0.352\\
    \midrule
    SNRE  &   \underline{9,452}  &  11.5569    &   0.0582   &  \textbf{0.948}\\
    \bottomrule
    \end{tabular}%
  \label{tab:complexity}%
\end{table}%

This subsection evaluate the model size and efficiency of different methods, in order to investigate whether the performance advantage of SNRE is obtained at the cost of substantially higher complexity. 
Specifically, on the PhishingWebsites dataset, we compare trainable parameters, training time, and single-sample inference time across methods, while also reporting the corresponding PNS, so as to jointly evaluate the trade-off between explanation quality and efficiency. 
Here, the training time reflects the overall cost of obtaining an explanation, the inference time reflects the efficiency of applying a learned explanation to an individual sample, and the model size characterizes the complexity of the method.

As shown in Table~\ref{tab:complexity}, SNRE achieves the best PNS of $0.948$ while maintaining a favorable efficiency--performance trade-off. 
Compared with high-performing baselines, SNRE uses only $9{,}452$ parameters, which is substantially smaller than RRL with $87{,}509$ parameters, and although slightly larger than DRNet with $6{,}240$ parameters, it delivers a much higher PNS. 
In terms of inference efficiency, SNRE requires only $0.0582$ ms per sample, making it faster than both RRL and DRNet. 
For training cost, SNRE is slower than lightweight rule-based methods such as IDS, FRL, and CORELS, but remains much faster than FLDT, while these faster methods generally lag far behind in PNS. 
Overall, SNRE is neither the smallest nor the fastest method in absolute terms, but it achieves the strongest sufficient-and-necessary explanation quality under a still moderate complexity budget, demonstrating a better overall cost-effectiveness.

\subsection{Case Study: Diagnosing Fine-Tuning-Induced Regional Prediction Drift} 
\label{sec:case_finetune_retention}

\begin{figure}[!t]
\centering
\includegraphics[width=\linewidth]{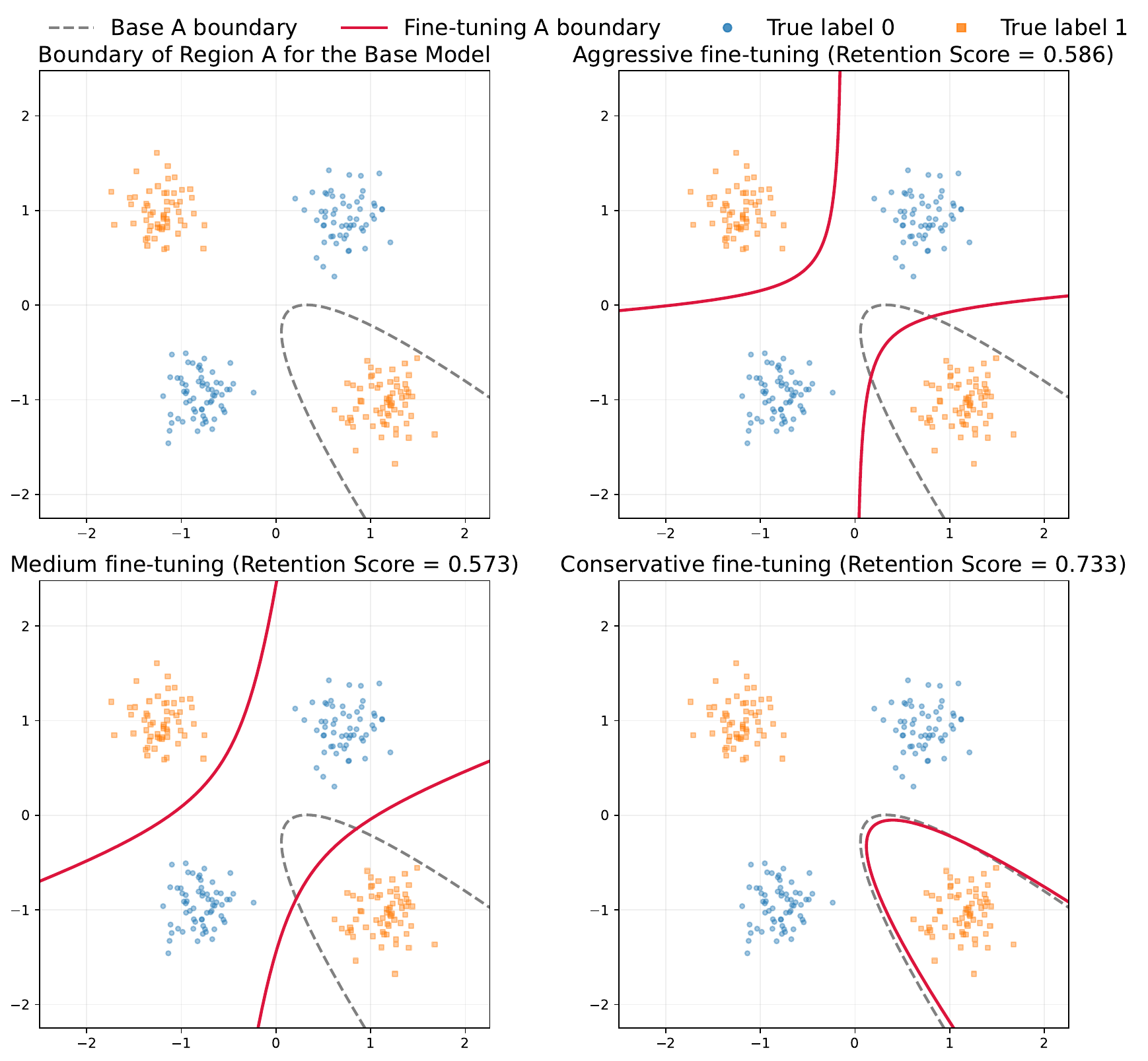}
\caption{Visualization of the retention of the dominant sufficiency--necessity explanation structure under fine-tuning on a synthetic dataset.}
\label{fig:case_ft}
\end{figure}

This case study investigates whether SNRE can quantify how the learned sufficient-and-necessary prediction regions change after fine-tuning. 
We define the retention score as
\begin{equation}
\mathrm{Retention}
=
\frac{1}{3}
\left(
\mathrm{IoU}_A+
\mathrm{IoU}_{\mathrm{mask}}+
\mathrm{IoU}_B
\right),
\label{eq:retention_score}
\end{equation}
where the three terms measure the overlap of the input region, feature mask and output region, 
before and after fine-tuning. 
A larger value indicates stronger preservation of the learned sufficient-and-necessary regional pattern.

To validate this metric, we visualize the learned input-region boundaries on a two-dimensional synthetic dataset. 
In Fig.~\ref{fig:case_ft}, the gray dashed curve denotes the base-model boundary, and the red solid curve denotes the fine-tuned one. 
The \emph{Conservative fine-tuning} case achieves the highest retention score ($0.733$) and remains visually closest to the base boundary, with only a mild local adjustment. 
By contrast, \emph{Aggressive fine-tuning} and \emph{Medium fine-tuning} obtain lower retention scores of $0.586$ and $0.573$, respectively, and exhibit more evident geometric deviations. 
This shows that the proposed retention score aligns well with visual intuition and provides a practical measure of fine-tuning-induced regional drift.

\subsection{Case Study: Active Learning via SNRE-Guided Sampling}
\begin{figure}[!t]
\centering
\includegraphics[width=\linewidth]{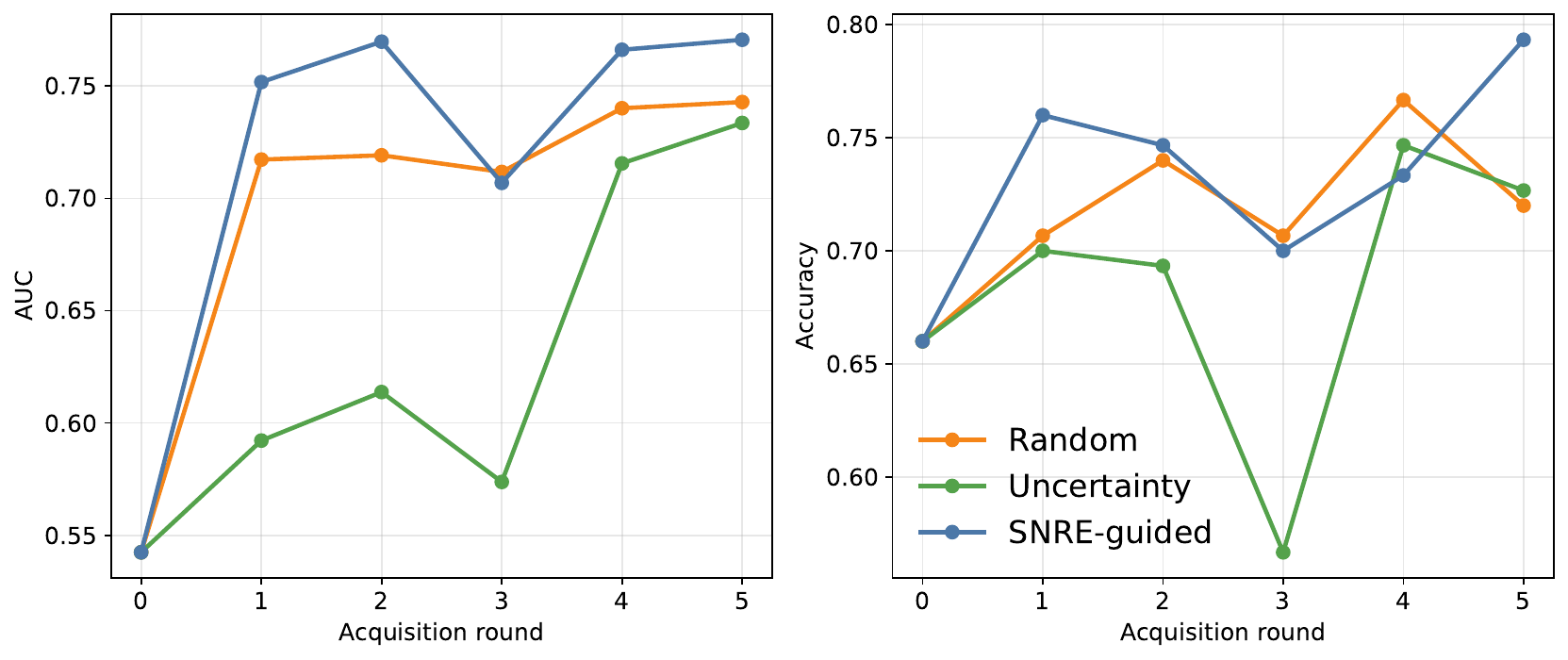}
\caption{Comparison of test AUC and accuracy on the German Credit dataset across active learning rounds for random sampling, uncertainty sampling, and the proposed SNRE-guided sampling.}
\label{fig:case_curves}
\end{figure}

\begin{figure}[!t]
\centering
\includegraphics[width=\linewidth]{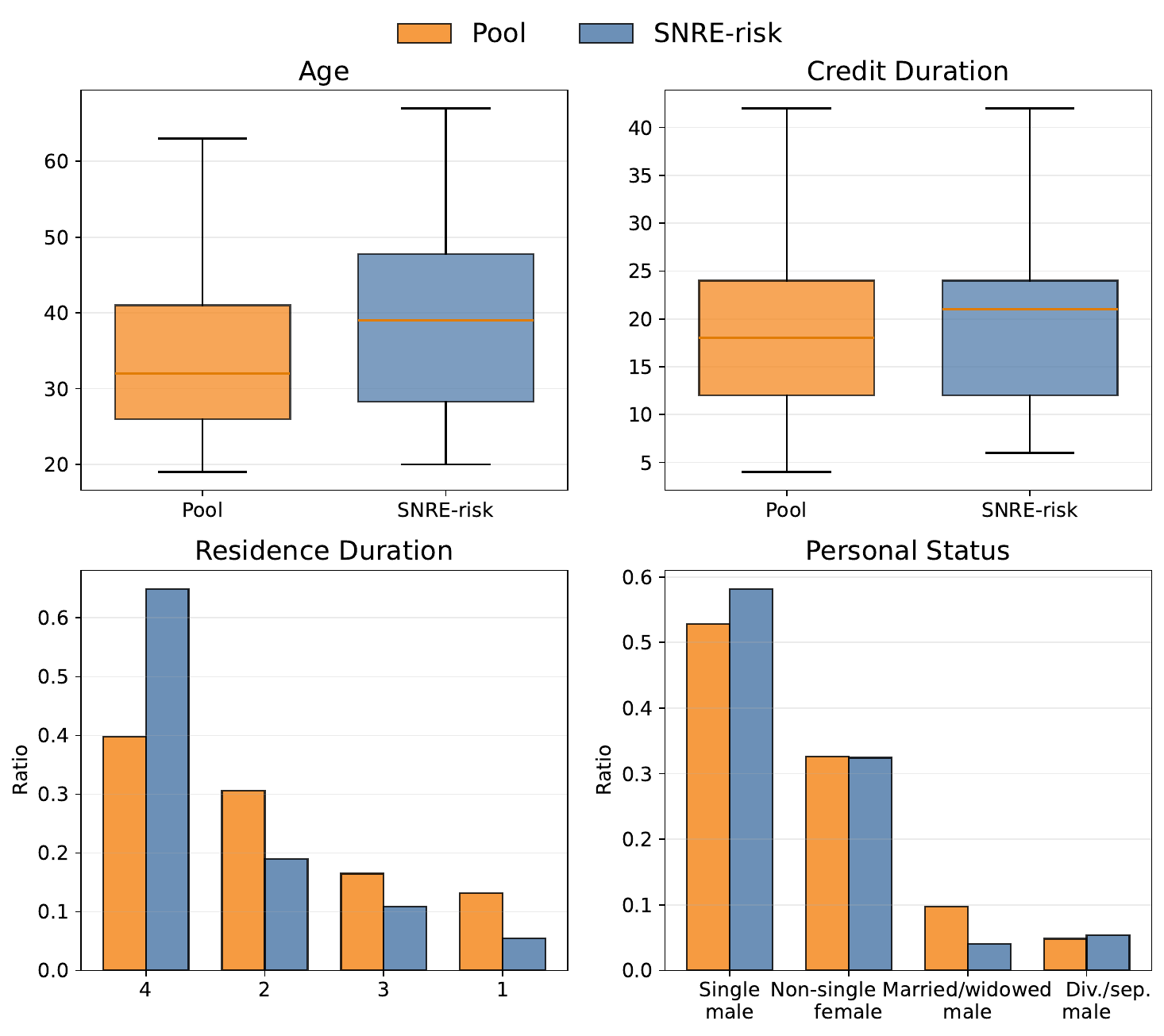}
\caption{Comparison of feature distributions on the German Credit dataset between the unlabeled samples and the samples acquired by SNRE-guided sampling.}
\label{fig:case_feature_grid}
\end{figure}

This case study investigates whether SNRE-guided sampling can select more informative unlabeled samples on the German Credit dataset\cite{settles2009active,gal2017deep}. 
The intuition is that, among unlabeled samples, those whose predictions already fall in the target output region while their inputs remain outside but near the learned sufficient-and-necessary region may be particularly informative for improving the model. 
Such samples suggest that the current prediction mechanism, especially its sufficient-and-necessary component captured by region $A$, may not yet fully account for a meaningful part of the input space. 
Labeling these samples may therefore help the model adjust this part of its prediction mechanism and improve subsequent predictive performance.

Based on this intuition, we heuristically assign each pool sample $x$ the acquisition score
\begin{equation}
s(x)=\sigma\!\left(\frac{p(x)-c_{\mathrm{low}}}{\tau_f}\right)
      \exp\!\left(-\frac{g_+(x)}{\tau_A}\right),
\end{equation}
where $p(x)$ is the predicted positive probability, $c_{\mathrm{low}}$ is the lower threshold of the target output region, and $g_+(x)=\max(g(x),0)$ measures how far $x$ lies outside the learned sufficient-and-necessary region $A$. 
Thus, larger scores correspond to samples whose predictions already fall in the target region and whose inputs lie just outside, but close to, the current sufficient-and-necessary component of the model.

Results show that SNRE-guided sampling consistently achieves higher and more stable test AUC and accuracy, with faster convergence (Fig.~\ref{fig:case_curves}). 
In contrast, random and uncertainty sampling exhibit larger performance fluctuations. 
This suggests that adding high-confidence samples near but outside the learned SNRE input region is more beneficial for subsequent model improvement than adding arbitrary or merely uncertain instances.

To characterize the samples selected by SNRE-guided sampling, we compare their feature distributions with those of the overall unlabeled pool, as shown in Fig.~\ref{fig:case_feature_grid}. 
The most distinctive attributes identified by distribution shift indicate that SNRE-guided sampling tends to select applicants who are older, have longer loan durations, have lived longer at their current residence, and are more often single males. 
These samples are not scattered outliers but form a coherent subgroup with consistent characteristics. 
In other words, the model has begun to assign high confidence to this subgroup, yet the learned SNRE input region $A$ does not stably cover it. 
Labeling such samples encourages the model to recalibrate its predictions on this subgroup, improving overall performance and expanding the explanatory region to capture the newly learned regularity.

\section{Conclusion}
In this paper, we proposed Sufficient and Necessary Regional Explanations (SNRE), a framework for explaining a trained predictor through input-output region pairs.
By formulating a region-level version of the Probability of Necessity and Sufficiency, SNRE provides a principled way to characterize when and only when a prediction behavior arises.
We introduced stochastic interventions to make region-level sufficiency and necessity well-defined and computable from finite data, and developed a differentiable objective for learning such regions end to end.
To balance expressiveness and interpretability, SNRE parameterizes regions with quadratic hypersurfaces and incorporates a learnable feature mask.
Experiments on real-world and synthetic datasets show that SNRE achieves strong sufficiency-necessity quality and robust explanation behavior.
Case studies further illustrate the practicality of the learned region pairs.

\bibliography{citation}
\bibliographystyle{IEEEtran}



\appendices

\section{Monte Carlo Estimation of Continuous PNS}
\label{app:pns_mc}
In this appendix, we provide a justification for the Monte Carlo estimator in \eqref{eq:pns_mc}. 
Recall that, for an input region $A\subseteq\mathcal D$ and an output region $B\subseteq\mathrm{Val}_{\mathbf Y}$, the definition of PNS over regions in continuous domains (Definition~\ref{def:cont_pns}) is
\begin{equation}
\mathrm{PNS}(A,B)
=
P\Big(
\mathbf Y_{\pi(\mathbf X;P^A)} \in B 
\;\wedge\;
\mathbf Y_{\pi(\mathbf X;P^{\bar A})} \in \bar B
\Big),
\end{equation}
where
\begin{equation}
\label{eq:cond_pns_appendix}
P^A(\mathbf X)=P^*(\mathbf X\mid \mathbf X\in A),
\quad
P^{\bar A}(\mathbf X)=P^*(\mathbf X\mid \mathbf X\in\bar A).
\end{equation}

Since the predictor $f$ is deterministic and the input variables $\mathbf X$ have no parents, 
each intervention fully determines the model output.
Moreover, the two stochastic intervention policies 
$\pi(\mathbf X;P^A)$ and $\pi(\mathbf X;P^{\bar A})$ are independently sampled. 
As a result, the joint probability factorizes as
\begin{equation}
\label{eq:pns_factorization}
\mathrm{PNS}(A,B)
=
P\big(\mathbf Y_{\pi(\mathbf X;P^A)} \in B\big)\,
P\big(\mathbf Y_{\pi(\mathbf X;P^{\bar A})} \in \bar B\big).
\end{equation}

Under a hard intervention $do(\mathbf X = \mathbf x)$, the model output is given by
\begin{equation}
\mathbf Y_{do(\mathbf X = \mathbf x)} = f(\mathbf x).
\end{equation}
Therefore, under the stochastic intervention $\pi(\mathbf X;P^A)$, we have
\begin{equation}
P\big(\mathbf Y_{\pi(\mathbf X;P^A)} \in B\big)
=
\mathbb E_{\mathbf X \sim P^*(\cdot \mid \mathbf X \in A)}
\big[
\mathbb I\big(f(\mathbf X) \in B\big)
\big],
\end{equation}
and analogously,
\begin{equation}
P\big(\mathbf Y_{\pi(\mathbf X;P^{\bar A})} \in \bar B\big)
=
\mathbb E_{\mathbf X \sim P^*(\cdot \mid \mathbf X \in \bar A)}
\big[
\mathbb I\big(f(\mathbf X) \in \bar B\big)
\big].
\end{equation}

Let 
$\{\mathbf x_i^{\mathrm{in}}\}_{i=1}^m 
\overset{\mathrm{i.i.d.}}{\sim} 
P^*(\mathbf X \mid \mathbf X \in A)$
and
$\{\mathbf x_i^{\mathrm{out}}\}_{i=1}^m 
\overset{\mathrm{i.i.d.}}{\sim} 
P^*(\mathbf X \mid \mathbf X \in \bar A)$.
We define the Monte Carlo estimator of $\mathrm{PNS}(A,B)$ as
\begin{equation}
\label{eq:pns_mc_appendix}
\widehat{\mathrm{PNS}}(A,B)
\!=\!
\left(\frac{1}{m}\!\sum_{i=1}^{m}\!\mathbb I\!\big[f(\mathbf x_i^{\mathrm{in}})\!\in\!B\big]\!\right)\!
\left(\frac{1}{m}\!\sum_{i=1}^{m}\!\mathbb I\!\big[f(\mathbf x_i^{\mathrm{out}})\in\!\bar B\big]\!\right)\!.
\end{equation}

By the law of large numbers, each empirical average converges almost surely to its corresponding conditional probability in \eqref{eq:cond_pns_appendix}.
Consequently,
\begin{equation}
\widehat{\mathrm{PNS}}(A,B)
\xrightarrow[m\to\infty]{\mathrm{a.s.}}
\mathrm{PNS}(A,B).
\end{equation}

\section{Derivation of the Empirical and Differentiable PNS Objective}
\label{app:pns_derivation}
This appendix derives the empirical objective used in \eqref{equ:final_obj}. 
Let the input variables be decomposed as $\mathbf X=(\mathbf I,\mathbf Z)$, where $\mathbf I$ denotes the intervened variables and $\mathbf Z$ is treated as the background variables to be marginalized. 
For an input region $A\subseteq \mathrm{Val}_{\mathbf I}$ and an output region $B\subseteq \mathrm{Val}_{\mathbf Y}$, the stochastic-intervention-based PNS can be written as
\begin{equation}
\label{eq:pns_expectation_app}
\begin{aligned}
\mathrm{PNS}(\mathbf I,A,B)
=
\mathbb E_{\mathbf z\sim P(\mathbf Z)}
\Big[
& P(\mathbf Y_{\pi(\mathbf I;P^A)}\in B\mid \mathbf z)\\
&\cdot
P(\mathbf Y_{\pi(\mathbf I;P^{\bar A})}\in \bar B\mid \mathbf z)
\Big],
\end{aligned}
\end{equation}
where $P^A(\mathbf I)=P(\mathbf I\mid \mathbf I\in A)$ and
$P^{\bar A}(\mathbf I)=P(\mathbf I\mid \mathbf I\in\bar A)$.
Since $f$ is deterministic,
\begin{equation}
P(\mathbf Y_{\pi(\mathbf I;P^A)}\in B\mid \mathbf z)
=
\mathbb E_{\mathbf i\sim P^A}
\big[\mathbb I(f(\mathbf i,\mathbf z)\in B)\big],
\end{equation}
and the complement term is defined analogously.

Given a dataset $\mathcal D=\{\mathbf x_n\}_{n=1}^N$, 
each sample is decomposed accordingly as $\mathbf x_n=(\mathbf i_n,\mathbf z_n)$.
The empirical intervention weights are
\begin{equation}
\omega^A(\mathbf x_j)
=
\frac{
\mathbb I(\mathbf i_j\in A)
}{
\sum_{\ell=1}^N
\mathbb I(\mathbf i_\ell\in A)
},
\qquad
\omega^{\bar A}(\mathbf x_j)
=
\frac{
\mathbb I(\mathbf i_j\in \bar A)
}{
\sum_{\ell=1}^N
\mathbb I(\mathbf i_\ell\in \bar A)
}.
\end{equation}
Then the empirical estimator is
\begin{equation}
\label{eq:hard_empirical_pns_app}
\begin{aligned}
\widehat{\mathrm{PNS}}(\mathbf I,A,B)
=
\frac{1}{N}\sum_{k=1}^N
\Big[
&\sum_{j=1}^N
\omega^A(\mathbf x_j)\,
\mathbb I\big(f(\mathbf i_j,\mathbf z_k)\in B\big)
\Big]\\
\cdot
\Big[
&\sum_{j=1}^N
\omega^{\bar A}(\mathbf x_j)\,
\mathbb I\big(f(\mathbf i_j,\mathbf z_k)\in \bar B\big)
\Big].
\end{aligned}
\end{equation}
This estimator averages over the background part $\mathbf z_k$ while drawing the intervened part $\mathbf i_j$ from the empirical in-region and out-of-region distributions.

Substituting the softening relaxations in \eqref{eq:soft_in}--\eqref{eq:soft_comp} and the mixing operation in \eqref{eq:mix_mask} into the hard empirical estimator \eqref{eq:hard_empirical_pns_app} yields the differentiable objective
\begin{equation}
\label{eq:soft_pns_app}
\begin{aligned}
&\widehat{\mathrm{PNS}}(A_{\theta, \mathbf{m}}, B_\phi)=\\
&\frac{1}{N}\sum_{k=1}^N\!\sum_{j=1}^N \omega^A_{\theta,\mathbf{m}}(\mathbf x_j)\, s^B_\phi\!\big(f(\tilde{\mathbf x}_{jk})\big)
\sum_{j=1}^N \omega^{\bar A}_{\theta, \mathbf{m}}(\mathbf x_j)\, s^{\bar B}_\phi\!\big(f(\tilde{\mathbf x}_{jk})\big),  
\end{aligned}
\end{equation}

\section{Details on Experiments}\label{app:detail_exp}
\subsection{Datasets}
We evaluate SNRE on four real-world tabular datasets and two synthetic datasets. 
\textbf{German Credit}\cite{hoffmanstatlog} contains personal and financial attributes of credit applicants. The label indicates whether the applicant is classified as good or bad credit.
\textbf{Phishing Websites}\cite{mohammad2014intelligent} consists of website and URL-related attributes. The label indicates whether a website is phishing or legitimate.
\textbf{Mushroom}\cite{mushroom_73} contains categorical attributes describing physical properties of mushrooms, such as cap shape, cap color, odor, gill size, stalk surface, veil type, and habitat. The label indicates whether a mushroom is edible or poisonous.
\textbf{Give Me Some Credit (GMSC)}\cite{GiveMeSomeCredit} contains financial and delinquency-related attributes. The label indicates whether an individual experiences serious delinquency within two years. 
\textbf{Synthetic Circles} is a two-dimensional synthetic dataset constructed from concentric circular structures. The label is determined by the constructed geometric region.
\textbf{Synthetic Blobs} is a two-dimensional synthetic dataset constructed from Gaussian clusters with controlled geometric structure. The label is generated according to the predefined cluster/region configuration.

\subsection{Baselines}
\subsubsection{Rule-based explanation baselines}
\textbf{IDS}~\cite{lakkaraju2016interpretable} learns interpretable decision sets, i.e., unordered sets of if--then rules optimized for accuracy, conciseness, and low rule overlap. 
\textbf{SBRL}~\cite{pmlr-v70-yang17h} learns scalable Bayesian rule lists by optimizing a posterior over ordered if--then rules constructed.  \textbf{FRL}~\cite{pmlr-v38-wang15a} learns falling rule lists, where the predicted probability is constrained to decrease monotonically along the ordered list. \textbf{DT}~\cite{loh2011classification} denotes a standard decision tree classifier trained by recursive feature splitting. The input event $A$ is obtained by taking the union of leaves that predict the target behavior. \textbf{FLDT}~\cite{good2023feature} denotes a feature-learning decision tree, which alternates between learning a sparse feature transformation and fitting a differentiable decision tree on the learned representations. \textbf{RRL}~\cite{wang2021scalable} denotes Rule-based Representation Learner, which learns non-fuzzy logical rules through a differentiable relaxation and gradient-based optimization. 
\textbf{CORELS}~\cite{angelino2018learning} learns certifiably optimal rule lists for categorical data by optimizing regularized empirical risk with an optimality certificate. 
\textbf{SamRuLe}~\cite{pellegrina2024scalable} is a sampling-based scalable rule-list learner that approximates optimal rule lists on large datasets with statistical guarantees. 
\textbf{MDL-RL}~\cite{proencca2020interpretable} denotes an MDL-based rule-list learner that selects compact probabilistic rule lists according to the minimum description length principle. 
\textbf{DRNet}~\cite{qiao2021learning} denotes a decision-rule network that learns interpretable decision rule sets using a neural architecture with logical rule components. 
\textbf{IMLI}~\cite{ghosh2022efficient} learns sparse propositional-logic classification rules through an iterative learning process that jointly optimizes accuracy and rule size.

\subsubsection{PNS-based explanation baselines}
\textbf{FANS}~\cite{chen2024feature} learns feature subsets by maximizing necessity-and-sufficiency effects under prediction perturbations. \textbf{LEWIS}~\cite{galhotra2021explaining} estimates global necessity and sufficiency scores through counterfactual interventions. \textbf{LENS}~\cite{watson2021local} identifies local minimal sufficient factors and aggregates them into global feature scores. 
\textbf{NSE4T}~\cite{balkir2022necessity} estimates local necessity and sufficiency scores through feature perturbations.
\textbf{CF2}~\cite{tan2022learning} combines factual and counterfactual reasoning to obtain necessary and sufficient explanations.
\textbf{NSEG}~\cite{cai2025probability} optimizes a lower bound of PNS to learn necessary and sufficient masks.

\subsection{Evaluation Metrics}
Different experiments use different evaluation metrics. We summarize them below.

\subsubsection{Sufficiency and necessity scores}
Let $\mathcal D_{\mathrm{eval}}=\{\mathbf x_k\}_{k=1}^{n}$ be the evaluation set.
Given the learned hard feature mask $\mathbf m\in\{0,1\}^{d}$, the input and output events are defined as
\begin{equation}
A=\{\mathbf x | g(\mathbf x;\mathbf m)\le 0\},
\qquad
B=\{\mathbf y | h(\mathbf y)\le 0\}.
\end{equation}
We denote the samples inside and outside the input region by
\begin{equation}
\mathcal D_A=\{\mathbf x_j\in\mathcal D_{\mathrm{eval}} | \mathbf x_j\in A\},
\qquad
\mathcal D_{\bar A}=\mathcal D_{\mathrm{eval}}\setminus \mathcal D_A .
\end{equation}

The empirical sufficiency score is computed as
\begin{equation}
\widehat{\mathrm{PS}}
=
\frac{1}{n}
\sum_{k=1}^{n}
\frac{1}{|\mathcal D_A|}
\sum_{\mathbf x_j\in\mathcal D_A}
\mathbb I\!\left[
f\!\left(
\mathbf m\odot \mathbf x_j+
(1-\mathbf m)\odot \mathbf x_k
\right)\in B
\right].
\end{equation}
Here, $\mathbf m\odot \mathbf x_j$ enforces the feature dimensions selected by $\mathbf{m}$ to take values from an in-region sample, while $(1-\mathbf m)\odot \mathbf x_k$, together with the empirical averaging over different samples $\mathbf x_k$, marginalizes the feature dimensions not selected by $\mathbf m$.


Similarly, the empirical necessity score is computed as
\begin{equation}
\widehat{\mathrm{PN}}
=
\frac{1}{n}
\sum_{k=1}^{n}
\frac{1}{|\mathcal D_{\bar A}|}
\sum_{\mathbf x_j\in\mathcal D_{\bar A}}
\mathbb I\!\left[
f\!\left(
\mathbf m\odot \mathbf x_j+
(1-\mathbf m)\odot \mathbf x_k
\right)\notin B
\right].
\end{equation}
It measures whether forcing the selected features to lie outside $A$ makes the model output fall outside $B$, while again marginalizing the unselected features through the evaluation samples.

\subsection{Implementation}



All predictors to be explained are three-layer MLPs with two hidden layers and one output layer. 
Adam is used as the optimizer. 
Each dataset is split into training, validation, and test sets, and predictor training is stopped early according to the validation loss, with a maximum of 200 epochs and a patience of 20 epochs.

For the regional-explanation experiments, the hidden dimension is 64, the batch size is 1024, and the data split is 60\%/20\%/20\%. 
SNRE uses a learning rate of 0.008 and an early-stopping patience of 25, with hyperparameters selected on the validation set.

\vfill

\end{document}